\documentclass[10pt,twocolumn,letterpaper]{article}

\usepackage[pagenumbers]{wacv} 

\usepackage{graphicx}
\usepackage{amsmath}
\usepackage{amssymb}
\usepackage{booktabs}
\usepackage{float}
\usepackage{titling}
\usepackage[pagebackref,breaklinks,colorlinks]{hyperref}

\usepackage[capitalize]{cleveref}
\crefname{section}{Sec.}{Secs.}
\Crefname{section}{Section}{Sections}
\Crefname{table}{Table}{Tables}
\crefname{table}{Tab.}{Tabs.}

\def\wacvPaperID{2637} 
\def\confName{WACV}
\def\confYear{2025}

\begin{document}

\title{LASERS: LAtent Space Encoding for Representations with Sparsity for Generative Modeling}

\author{Xin Li\\
Rutgers, the State University of New Jersey\\
New Brunswick, NJ 08901-8554\\
{\tt\small xl598@scarletmail.rutgers.edu}
\and
Anand D. Sarwate\\
Rutgers, the State University of New Jersey\\
New Brunswick, NJ 08901-8554\\
{\tt\small ads221@soe.rutgers.edu}
}
\maketitle

\begin{abstract}
   Learning compact and meaningful latent space representations has been shown to be very useful in generative modeling tasks for visual data. One particular example is applying Vector Quantization (VQ) in variational autoencoders (VQ-VAEs, VQ-GANs, etc.), which has demonstrated state-of-the-art performance in many modern generative modeling applications. Quantizing the latent space has been justified by the assumption that the data themselves are inherently discrete in the latent space (like pixel values). In this paper, we propose an alternative representation of the latent space by relaxing the structural assumption than the VQ formulation. Specifically, we assume the latent space can be approximated by a union of subspaces model corresponding to a dictionary-based representation under a sparsity constraint. The dictionary is learned/updated during the training process. We apply this approach to look at two models: Dictionary Learning Variational Autoencoders (DL-VAEs) and DL-VAEs with Generative Adversarial Networks (DL-GANs). We show empirically that our more latent space is more expressive and has leads to better representations than the VQ approach in terms of reconstruction quality at the expense of a small computational overhead for the latent space computation. Our results thus suggest that the true benefit of the VQ approach might not be from discretization of the latent space, but rather the lossy compression of the latent space. We confirm this hypothesis by showing that our sparse representations also address the codebook collapse issue as found common in VQ-family models.
\end{abstract}

\section{Introduction}
\label{sec:intro}

Extracting meaningful latent representations has shown its impressive impact in various generative modeling applications and efficiently structuring the latent space is thus of continuing interest in machine learning research. Autoencoder-based approaches implicitly assume that the latent distribution falls into exponential family distributions~\cite{vae}. Without additional structural assumptions, variational autoencoders (VAEs) can suffer from ``mode collapse,'' which motivated the development of Vector Quantized Variational Autoencoders (VQ-VAEs)~\cite{vqvae, vqvae2}. This approach assumes (or enforces) a latent space that is well approximated by a Voronoi tesselation induced by quantization points (also called the ``codebook'').  This discretization of the latent space has seen many successes in generative modeling tasks~\cite{vqgan, ldm, improved-vqgan}, and the discrete codebook generated from the training phase can also help with transformer training~\cite{vqgan, attention, gpt}.

Vector quantization structures the latent space (the aforementioned Voronoi tesselation) and which we is a compression algorithm for the embedding vectors since all that needs to be stored is the index of the codebook. However, associating every embedding vector to a single codebook vector is an overly strict structural constraint. One proffered justification for the discretization is that in many applications the data of interest is already discretized into finite-bit representations (such as RGB images), or is itself categorical (such as text).  In this paper we will instead think of the problem as one of lossy compression of the embedding vectors. This opens the door to many different forms of lossy compression/approximation. Essentially, we can think of the problem as representation learning on the embedding vectors themselves by putting another encoder/decoder pair between the encoder/decoder of the original VAE.

We contend that compression is more important than discretization and explore an alternative compression strategy using sparse dictionary learning~\cite{cs}. Specifically, we learn a dictionary that can represent the embedding vectors as sparse linear combinations of atoms from an underlying dictionary. This structural assumption is a more relaxed version as compared to Vector Quantization since multiple codebook (dictionary) elements can be used to represent a single embedding vector. As our experiments show, this yields a much more expressive latent space. A secondary benefit over quantization is that the dictionary representation is more resistant to ``codebook collapse,'' which is a common issue in modern VQ-family model training with fixed-length codebooks~\cite{edvae}. 
Geometrically, we engineer the latent space to be structured as a \emph{union of subspaces}~\cite{cs}. 
In the next section, we will briefly review prior arts in VQ-family models, sparse coding and dictionary learning research, \etc

In summary, our major contributions are:
\begin{itemize}
    \item Modeling the latent space structure issue as a compression problem to design the Dictionary Learning Variational Autoencoder model (DL-VAE) and the DL-VAE with Generative Adversarial Networks (DL-GAN) model. 
    \item Designing effective training algorithms for these models to show how we can learn the latent space model with even more representation power.
    \item Demonstrating that our proposed approach improves the reconstruction quality using common image datasets such as MNIST~\cite{mnist}, CIFAR10~\cite{cifar10}, Oxford Flowers~\cite{oxford-flowers}, and FFHQ~\cite{ffhq}, \etc
    \item Experimental validation of the approach in terms of addressing the codebook collapse issue as seen in VQ-family models.
    \item In the supplementary material we show how our model can be applied to downstream generative modeling tasks such as inpainting, super-resolution and generation with diffusion models, we have also included the ablation studies in the supplementary materials.
\end{itemize}

The primary objective of this paper is to open the door to a wider range of approaches for improving latent space representations, rather than achieving state-of-the-art performance. We contend that imposing structural constraints on the latent space seems to be the primary benefit of the VQ approach, rather than discretization. This suggests that many kinds of lossy compression technique could be beneficial in these applications.

\medskip

\section{Related Works}
\label{sec:related}

The idea of discretizing the latent space we study here was initially proposed in the context of variational autoencoders~\cite{vae} and has shown great success in various generation tasks and produced state-of-the-art results. Moreover, sparse dictionary learning and its application in autoencoders have been of interest over the past few years.

\noindent \textbf{VQ-VAE and VQ-VAE2.} Variational autoencoders with a discretization bottleneck~\cite{vqvae, vqvae2} are models which are better at data compression and can avoid the posterior collapse issue~\cite{posterior-collapse} as seen in many other variational autoencoder-based models. VQ-VAE2~\cite{vqvae2} is an extension over VQ-VAE with two layers of quantization, which yields even higher fidelity results.

\noindent \textbf{DALL-E Based Models.} the DALL-E~\cite{dalle} family models as proposed by OpenAI replace the vector quantization with Gumbel-Softmax for categorical reparametrization~\cite{gumbel-softmax} so that the gradient can flow through the vector quantization bottleneck. 

\noindent \textbf{VQ-GAN.} This variant on the VQ-VAE yields higher generation quality with perceptual loss and a generative adversarial network (GAN)~\cite{vqgan}. The learned codebook used in the VQ stage has also been shown to be helpful for transformer training~\cite{vqgan, attention, gpt}.

\noindent \textbf{Stable Diffusion.} The VQ-GAN can also be used to obtain a high-quality latent space for high-resolution image synthesis~\cite{ldm}. This provides an efficient extension on powerful but computationally demanding diffusion models~\cite{diffusion, ddpm, beat-gans}.

\noindent \textbf{Sparse Coding and Dictionary Learning.} Many approaches have been proposed for the dictionary learning problem such as methods based on LASSO regression~\cite{online-dl}, Stochastic Gradient Descent~\cite{sgd-dl}, and k-SVD~\cite{ksvd, efficient-ksvd}. Among the dictionary learning algorithms, most approaches are often implemented in a two stage fashion, with a sparse coding stage of implemented using either greedy pursuit algorithms, such as Matching Pursuits, Orthogonal Matching Pursuits (OMP)~\cite{cs} and its more efficient variant Batch Orthogonal Matching Pursuits (Batch-OMP)~\cite{efficient-ksvd}, or thresholding algorithms, such as Iterative Hard Threshold (IHT)~\cite{iht}, Iterative Shrinkage and Thresholding Algorithm (ISTA)~\cite{ista} and its variants, and a learnable formulation such as Learnable-ISTA (LISTA)~\cite{lista}, \etc.

\noindent \textbf{Other VAEs with Sparse Coding.} Earlier attempts on applying sparse dictionary learning on VAE models involve learning the representation as the encoder output with a fixed dictionary based on Discrete Cosine Transform (DCT)~\cite{sparsity-promoting}, which does not yield a dictionary embedded with the rich semantic information for other potential applications, the model introduced in~\cite{sparsity-promoting} also focuses on audio signal processing and not image data. Another more recent method proposed in~\cite{sc-vae}, while taking a similar format as our model by taking the encoder output directly as the sparse codes through a dictionary learning bottleneck, however it still relies on a predetermined orthogonal dictionary and does not discuss its applicability with the discriminator network for sharper outputs. The Sparse-VAE~\cite{sparse-vae}, though similarly enforcing a sparsity constraint, but rather it focuses on a totally different problem where the goal is to reduce the latent space feature dimension using sparsity, and does not learn a semantically rich dictionary.

\section{Dictionary Learning of Latent Spaces}
\label{sec:structuring-latents}

\begin{figure}[H]
    \centering
    \includegraphics[width=\linewidth]{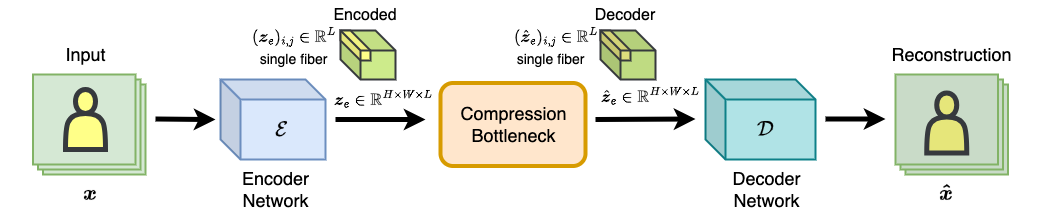}
    \caption{Architecture of a generic autoencoder model with the compression bottleneck.}
    \label{fig:generic}
\end{figure}

To set up a generic framework to structure the latent space, as with standard VAEs, we can structure the autoencoder class models with an encoder, a decoder, and a compression bottleneck as in Figure \ref{fig:generic} follows~\cite{vae},

Note that for the most basic VAE model, the compression bottleneck as specified in the diagram above is essentially the identity in theory, and thus $\boldsymbol{z}_e = \hat{\boldsymbol{z}}_e$, although in practice it often involves a reparameterization procedure for latent space sampling~\cite{vae}.

\subsection{Prior work: Vector Quantization}
\label{subsec:vq}

An earlier successful attempt in structuring the latent space following this generic framework is to use the Vector Quantization~\cite{CoverThomas}, yielding the model of Vector-Quantized Variational Autoencoders (VQ-VAE)~\cite{vqvae, vqvae2}, where we assume a discrete latent space on top of the traditional Variational Autoencoder (VAE) models~\cite{vae}. The overall architecture is the same as in Figure \ref{fig:generic}, the a codebook matrix $\boldsymbol{E} \in \reals^{K \times L}$ inside the compression bottleneck. Specifically, as with standard VAEs, given the input data $\boldsymbol{x}$ and the latent space representation $\boldsymbol{z}$, we define the basic building blocks of the VQ-VAE model as follows,
\begin{itemize}
    \item An encoder network $\mathcal{E}$ which compresses the input $\boldsymbol{x}$ into the latent representation $\boldsymbol{z}$ by learning the parameters of a posterior distribution $q(\boldsymbol{z} | \boldsymbol{x})$;
    \item A prior distribution of the latent representation $p(\boldsymbol{z})$;
    \item A decoder network $\mathcal{D}$ that can recover $\boldsymbol{x}$ from the latent representation $\boldsymbol{z}$.
\end{itemize}
Then we denote the latent representation tensor generated from the encoder network $\mathcal{E}$ as $\boldsymbol{z}_e \in \reals^{H \times W \times L}$, where $H$ and $W$ represent the height and width dimensions of the latent space tensor, $L$ represents the feature dimension of the latent space. Here we assume in VQ-VAE both the prior and posterior distributions are categorical. In particular, we define the set of codebook vectors $\mathfrak{E} = \{\boldsymbol{e}_k \in \reals^L | k \in [K] \}$, where $K$ is the number of embedding vectors, $[B]$ represents the set of natural numbers upper bounded by an integer $\boldsymbol{B} \in \mathbb{N}^\dagger$, \ie, $[B] = \{1, 2, \cdots, B\}$. We then denote the latent representation $\boldsymbol{z} \in \reals^{H \times W}$ computed by the nearest-neighbor index mapping between the latent feature vectors and the shared codebook/embedding matrix $\boldsymbol{E} \in \reals^{K \times L}$, formed by stacking the codebook vectors from the set $\mathfrak{E}$, where each entry of $\boldsymbol{z}$ represents the index of the corresponding embedding vector in the codebook. The derived categorical distribution for $(i, j)$-th fiber where $i \in [H], j \in [W]$ is as the following,
\begin{equation}
    q(z_{ij} = k | \boldsymbol{x}) = \begin{cases}
        1 &\text{ if } k = \arg \min_{k'} \| (\boldsymbol{z}_e)_{ij} - \boldsymbol{e}_{k'} \|_2;\\
        0 &\text{ otherwise.}
    \end{cases}
\end{equation}
Finally, we denote the quantized latent representation tensor as $\hat{\boldsymbol{z}}_e \in \reals^{H \times W \times L}$, constructed by stacking the embedding vectors as indexed by the latent code $\boldsymbol{z}$,
\begin{equation}
    (\hat{\boldsymbol{z}}_e)_{ij} = \boldsymbol{e}_k \text{ where } z_{ij} = k.
\end{equation}
The quantized latents are then passed in to the decoder network $\mathcal{D}$ that learns the distribution $p(\boldsymbol{x} | \hat{\boldsymbol{z}}_e)$ to generate the reconstruction. The training objective of the VQ-VAE model can be expressed as a modified version of the ELBO loss~\cite{elbo, vqvae, vqgan},
\begin{align}
    \ell_{\text{VQ-VAE}}(\vec{x}, \hat{\vec{x}}) = &\ell_{\text{recon}} (\vec{x}, \hat{\vec{x}}) + \norm{\mathtt{sg} [\vec{z}_e] - \hat{\vec{z}}_e}_2^2 +\nonumber\\
    &\quad \beta \norm{\vec{z}_e - \mathtt{sg}[\hat{\vec{z}}_e]}_2^2, \label{eq:vqvae-obj}
\end{align}
where,
\begin{equation}
    \hat{\boldsymbol{z}}_e = \texttt{one-hot}[\boldsymbol{z}] \cdot \boldsymbol{E} \in \reals^{H \times W \times L}. \label{eq:vq-recon}
\end{equation}
The \texttt{one-hot} operator above performs one-hot encoding on the latent code $\boldsymbol{z} \in \reals^{H \times W}$ and converts it into an $H \times W \times L$ tensor, where each fiber along the channel dimension,
\begin{equation}
    (\texttt{one-hot}[\boldsymbol{z}])_{ij} = \mathds{1}_{\boldsymbol{z}} (k) = \begin{cases}
        1 & k = z,\\
        0 & \text{otherwise}.
    \end{cases} \in \reals^K.
\end{equation}
The $\mathtt{sg}[\cdot]$ operator in Equation \eqref{eq:vqvae-obj} denotes the stop gradient operation, which essentially detaches the operand tensor from the computation graph as in most modern deep learning frameworks based on automatic differentiation engines~\cite{pytorch}. The $\ell_{\text{recon}}$ in Equation \ref{eq:dlvae-obj} is a generic loss function to measure reconstruction quality, where in our experiments we use a combination of the $\ell_2$ loss and the perceptual loss $\ell_{\text{perceptual}}$~\cite{lpips} as suggested in~\cite{vqgan} for better results in perceptual quality,
\begin{align}
    \mathcal{L}_{\text{recon}} (\vec{x}, \hat{\vec{x}}) = \eta \| \vec{x}, \hat{\vec{x}} \|_2^2 + (1 - \eta) \cdot \ell_{\text{perceptual}}(\vec{x}, \hat{\vec{x}}),
\end{align}
with $\eta$ being a weighting factor that determines the ratio of $\ell_2$ loss and perceptual loss in the overall reconstruction loss function, in practice we find out that setting $\eta = 0.5$ yields satisfying results. The second and third term in Equation \eqref{eq:vqvae-obj} optimize the encoder output and the codebook alternatively by fixing either one term in the $\ell_2$ loss, the $\beta$ term associated with the third loss term in Equation \eqref{eq:vqvae-obj} is called the \textit{commitment loss}~\cite{vqvae}, which is used to constrain the third term so that it does not grow too large for the overall objective function to converge. A hard coded value of $0.25$ for $\beta$ has been shown to be effective in our experiments in align with~\cite{vqvae}. 

Note also that the Vector Quantization procedure involves a nearest-neighbor mapping which is not differentiable. In practice we apply a straight-through gradient estimator~\cite{straight-through-estimator} to copy the gradients from the encoder output directly to the compression bottleneck output: 
\begin{equation}
    \hat{\boldsymbol{z}}_e = \boldsymbol{z}_e + \mathtt{sg}[\hat{\boldsymbol{z}}_e - \boldsymbol{z}_e].
\end{equation}
As with standard VAEs, the quantization/discretization process happens in between the encoder and the decoder can also be thought of as the compression bottleneck of the model, which is crucial to the model performance in terms of data representation. 



Note that this formulation of the latent space can also be modeled into a probability distribution, which makes it viable to be thought of as a VAE~\cite{vae}. Specifically, the sparsity of the coefficients for combining the dictionary atoms is modeled with a Mixture-of-Gaussian distribution~\cite{sc-mixture} where one Gaussian captures nonactive coefficients with a small-variance distribution centered at zero, and one or more other Gaussian capture active coefficients with a large-variance distribution. More concretely, for the latent representation problem $\boldsymbol{z}_e = \boldsymbol{\gamma}\boldsymbol{D}, \mathtt{supp}\{ \boldsymbol{\gamma} \} = S$, the prior probability distribution over the coefficients $\boldsymbol{\gamma}_k, k = 1, 2, \cdots, K$, is factorial, with the distribution over each coefficient $\boldsymbol{\gamma}_k$ modeled as a Mixture-of-Gaussians distribution with two Gaussians. Then we have a set of binary state variables $a_k \in \{ 0, 1 \}$ to determine which Gaussian is used to describe the coefficients.

The total prior over both sets of variables $\boldsymbol{\gamma}$ and $\boldsymbol{a}$, is of the form,
\begin{equation}
    q(\boldsymbol{\gamma}, \boldsymbol{a}) = \prod_{k = 1}^K q(\boldsymbol{\gamma}_k | a_k) q(a_k),
\end{equation}
where $q(a_i)$ determines the probability of being in the active or inactive states, and $q(\boldsymbol{\gamma}_k | s_k)$ is a Gaussian distribution whose mean and variance is determined by the current state $s_k$.

The total probability for the latent representation can then be modeled by,
\begin{equation}
    q(\boldsymbol{z}_e ; \boldsymbol{\theta}) = \sum_{\boldsymbol{a}} q(\boldsymbol{s} ; \boldsymbol{\theta}) \int q(\boldsymbol{z}_e | \boldsymbol{\gamma}, \boldsymbol{\theta}) q(\boldsymbol{\gamma} | \boldsymbol{a} ; \boldsymbol{\theta}) d \boldsymbol{\gamma}.
\end{equation}

\subsection{Structuring with Dictionary Learning}

Vector Quantization compresses the latent space into a discrete set of codebook vectors. Our proposal is to use a different form of lossy compression by putting representation learning inside the VAE (essentially learning a better representation of the latent representations). Specifically, we apply sparse dictionary learning~\cite{cs} that allows for richer representations/structuring of the embedding space than the vector quantization. In dictionary-based representations, each embedding vector is approximated by a sparse linear combination of embedding vectors (dictionary atoms). This allows multiple atoms to represent one embedding vector and allows the contribution of each atom to vary its level of contribution by its coefficient. We show in our experimental work that this approach gives much stronger representing power with a modest computational overhead.

Specifically, given the set of dictionary atoms $\mathfrak{D} = \{\boldsymbol{d}_k \in \reals^L | k = 1, 2, \cdots, K \}$, we consider a sparse set of dictionary atoms whose the linear combination represents that individual latent feature vector, where we define number of dictionary atoms used for each latent feature vector as the sparsity level as $S \in \mathbb{N}^\dagger$ (note that when $S = 1$, we essentially get the VQ-VAE). Then we have the latent representation becomes a tensor $\boldsymbol{z} \in \mathbb{R}^{H \times W \times S}$, where each vector $\boldsymbol{z}_{ij} \in \mathbb{R}^S$ represents an indexing vector to select a sparse set of $S$ dictionary atoms. The compressed latent representation $\boldsymbol{z}$ directly corresponds to the sparse representation coefficient matrix, also known as the sparse codes, denoted as $\boldsymbol{\gamma} \in \mathbb{R}^{H \times W \times K}$, with each single fiber $\boldsymbol{\gamma}_{ij} \in \reals^K,  i \in [H], j \in [W]$,
\begin{align}
    \boldsymbol{\gamma}_{ij}, \boldsymbol{D} = \arg &\min_{\Tilde{\boldsymbol{\gamma}}_{ij}, \Tilde{\boldsymbol{D}}} \| (\boldsymbol{z}_e)_{ij} - \Tilde{\boldsymbol{\gamma}}_{ij} \Tilde{\boldsymbol{D}}\|_2,\\
    &\quad \textit{subject to}\quad \mathrm{supp}\{ \boldsymbol{\gamma}_{ij} \} = S,
\end{align}
where the $\boldsymbol{D} \in \mathbb{R}^{K \times L}$ is the dictionary matrix constructed from stacking the dictionary atom vectors in the dictionary set $\mathfrak{D}$, $\mathrm{supp}(\boldsymbol{\gamma}_{ij})$ denotes the support of the sparse code vector $\boldsymbol{\gamma}_{ij}$, which is defined the $\ell_0$ norm of the vector, \ie, $\mathrm{supp}(\boldsymbol{\gamma}_{ij}) = \| \boldsymbol{\gamma}_{ij} \|_0$. We name the derived model as Dictionary Learning Variational Autoencoder (DL-VAE). The high-level working mechanism of the DL-VAE model still follows the generic framework as depicted in Figure \ref{fig:generic}, with the compression bottleneck block containing a dictionary matrix $\boldsymbol{D} \in \reals^{K \times L}$.

\begin{figure*}[hbt!]
  \centering
  \begin{subfigure}{0.4\linewidth}
    \includegraphics[width=\textwidth]{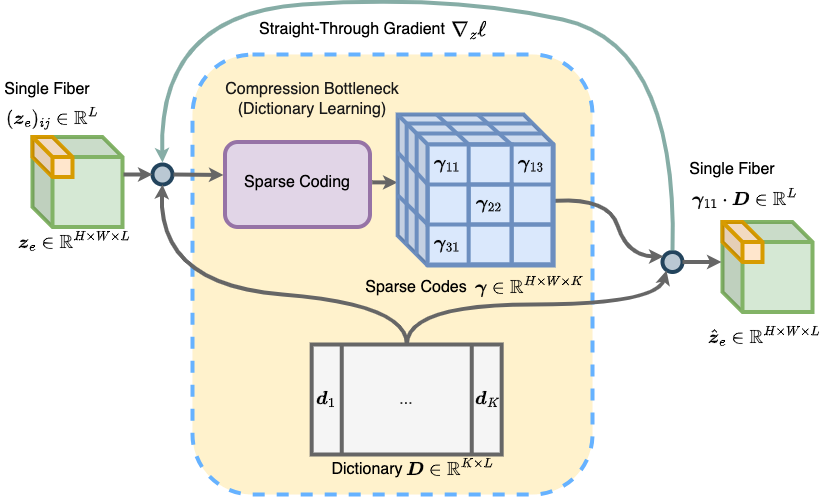}
    \caption{Dictionary learning bottleneck.}
    \label{fig:dl-internal}
  \end{subfigure}
  \hspace{2em}
  \begin{subfigure}{0.45\linewidth}
    \includegraphics[width=\textwidth]{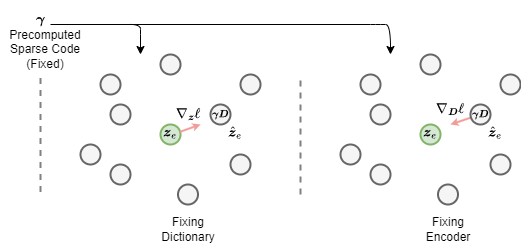}
    \caption{Dictionary atoms and encoder outputs during training.}
    \label{fig:dl-losses}
  \end{subfigure}
  \caption{(a) The internal working mechanism of the Dictionary Learning Compression Bottleneck. Note that here each fiber of the latent sparse codes is of $K$ length, however, due to the sparse structure assumption, we can always represent the sparse codes in a sparse data structure, such as the Sparse COO, CSR, \etc, data formats~\cite{cs, pytorch}, in which the fibers can be reduced to $S \ll K$ length; (b) How the encoder outputs and the dictionary atoms move towards each other during training.}
  \label{fig:dl-mechanism}
\end{figure*}

Note that in conventional dictionary learning setting we expect to solve an overcomplete problem~\cite{cs}, where we assume the feature dimension $L$ should be no greater than the number of dictionary atoms $K$. Furthermore, as the input image resolution gets higher, a larger latent space is required to maintain the representation quality, making it challenging to model the latent space distribution for the whole image. Thus in practice, we typically adopt a patch-level coding~\cite{emergence, overcomplete, sensory}, where we first unfold the latent space representation as non-overlapping patches \footnote{While conventionally people use overlapping image patches to make sure patches are related to each other, in our scenario since we already have a convolutional encoder and decoder network that make sure each latent feature vector covers an overlapping region of the original image, thus we do not need to create overlapping patches in the latent space, saving some additional compute time.} of dimension $P_H \times P_W \times L$ such that $P_H \times P_W \leq K$ before we perform the dictionary learning procedure to reconstruct the patches, where $P_H, P_W$ represents the respective height and width dimension of the patches or unfolding kernel size in implementation, then we reconstruct the latent space representation by folding back the reconstructed patches. The model shares the same versatility of choices of encoder and decoder architectures as the VQ family models (VQ-VAE, VQ-GAN, \etc)~\cite{vqvae, vqvae2, vqgan}. As for the compression bottleneck based on dictionary learning, we first assume the dictionary is prefixed and perform a sparse coding stage~\cite{ista, lista, omp, efficient-ksvd}, and then learn the dictionary via common momentum-based stochastic gradient optimizers~\cite{rmsprop, adam} while fixing the learned sparse codes. The diagram of the internal mechanism of the dictionary learning compression bottleneck is shown in Figure 3 \subref{fig:dl-internal}.

\subsection{Selecting the Best Dictionary Atoms}

To select the best sparse set of dictionary atoms and coefficients for the linear combinations, we first assume the learned dictionary is fixed, then perform the sparse coding procedure in the dictionary learning bottleneck on the encoded input. While there exist many methods for sparse coding, we adopted Batch Orthogonal Matching Pursuit (Batch-OMP)~\cite{efficient-ksvd} to compute the sparse codes for dictionary atom combination, a greedy approach that selects the best matching dictionary atoms with target sparsity and precision, which has demonstrated better convergence property than thresholding based methods~\cite{cs} and is easier to be separated from the dictionary training as compared to other learnable procedures such as LISTA~\cite{lista}. Specifically, we keep selecting the dictionary atoms with maximum correlation to the input signals and then perform a progressive Cholesky procedure~\cite{efficient-ksvd} to solve for the sparse codes.

After the sparse coding procedure, we compute the latent space reconstruction $\hat{\boldsymbol{z}}_e$ by multiplying the sparse codes and the dictionary matrix,
\begin{equation}
    (\hat{\boldsymbol{z}}_e)_{ij} = \boldsymbol{\gamma}_{ij} \boldsymbol{D},\: i \in [H], j \in [W].
\end{equation}
Then pass it through the decoder to get the reconstruction $\hat{\boldsymbol{x}}$, the quality of the reconstruction is then measured by common image reconstruction metrics such as the MSE, PSNR, SSIM and deep learning based metrics such as LPIPS~\cite{lpips} and the FLIP metric~\cite{flip}, etc.

Note that during training, given an input tensor $\boldsymbol{x}$, both the sparse codes and the dictionary loss are jointly optimized as in the VQ-VAE model, with the combined objective for input $\hat{\boldsymbol{x}}$,
\begin{align}
    &\ell_{\text{DL-VAE}}(\vec{x}, \hat{\vec{x}}) = \ell_{\text{recon}} (\vec{x}, \hat{\vec{x}}) + \norm{\mathtt{sg} [\vec{z}_e] - \hat{\vec{z}}_e}_2^2 +\nonumber\\
    &\qquad\qquad\qquad\qquad\qquad\qquad \beta \norm{\vec{z}_e - \mathtt{sg}[\hat{\vec{z}}_e]}_2^2, \nonumber\\
    &\hat{\boldsymbol{z}}_e = \boldsymbol{\gamma} \cdot \boldsymbol{D} \in \reals^{H \times W \times L}. \label{eq:dlvae-obj}
\end{align}
Note that here the sparse codes $\boldsymbol{\gamma}$ are calculated via the sparse coding procedure described above, which involves a non-differentiable $\argmax$ operation, thus following the similar idea adopted in VQ-VAE training, we apply a straight-through gradient estimator to make sure the gradient backpropagates through the entire computational graph approximately. By incorporating the dictionary-related loss term (Equation \ref{eq:dlvae-obj}) in the objective, the model will perform an implicit online dictionary update~\cite{online-dl, noodl} with modern momentum-based stochastic optimization methods~\cite{rmsprop, adam}. Geometrically speaking, the dictionary atoms in combination with the precomputed sparse codes will move towards the latent feature vectors from the encoder output (as shown in Figure 3 \subref{fig:dl-internal}). Note that here we can also use a learning rate free rule to update the dictionary in an online manner as proposed by~\cite{online-dl}, but in practice it proves to be slow as the number of dictionary atoms increase and does not yield a superior performance. Also here we do not use the traditional dictionary learning algorithms as the datasets we use are rather large and thus not fit for traditional dictionary learning algorithms where we expect to use the entire dataset~\cite{cs, ksvd, efficient-ksvd}.

\subsection{Deblurring with the Discriminator Network}

Following the same idea as in~\cite{vqgan}, the DL-VAE model can also be enhanced by appending a discriminator network, \ie, the PatchGAN~\cite{patchgan} to produce even sharper output. Specifically, we introduce a patch-based discriminator $\mathfrak{D}$ that will try to distinguish the original image ($\vec{x}$) and the reconstructed image ($\hat{\vec{x}}$) using the GAN objective,
\begin{align}
    \ell_{\mathrm{GAN}}(\{\mathcal{E}, \mathcal{G}, \mathcal{Z}\}, \mathfrak{D}) = \left[ \log \mathfrak{D}(\vec{x}) + \log (1 - \mathfrak{D}(\hat{\vec{x}}) \right],
\end{align}
where $\mathcal{G}$ denotes the generic generative network, which is just the decoder network $\mathcal{D}$ in our VAE models. The we have the complete objective for the DL-GAN model $\mathfrak{Q}^\ast = \{ \mathcal{E}^\ast, \mathcal{G}^\ast, \mathcal{Z}^\ast \}$ is the solution to the minimax game,
\begin{align}
    \mathfrak{Q}^\ast = \underset{\mathcal{E}, \mathcal{G}, \mathcal{Z}}{\arg \min} \:\underset{\mathfrak{D}}{\max} &\mathbb{E}_{p_{\set{D}} (\vec{x})} \big[ \ell_{\mathrm{DL-VAE}}(\mathcal{E}, \mathcal{G}, \mathfrak{\mathcal{Z}}) +\nonumber\\
    &\qquad \lambda \ell_{\mathrm{GAN}}(\{\mathcal{E}, \mathcal{G}, \mathcal{Z}\}, \mathfrak{D}) \big],
\end{align}
where the $\lambda$ is the adaptive weight computed as follows,
\begin{align}
    \lambda = \frac{\nabla_{\mathcal{G}_\ell} [\ell_{\text{recon}}]}{\nabla_{\mathcal{G}_\ell}[\ell_{\mathrm{GAN}}] + \delta}
\end{align}
where $\nabla_{\mathfrak{G}_L}[\cdot]$ denotes the corresponding gradient with respect to the last layer of the decoder, and $\delta$ is small constant for numerical stability, where we pick $\delta = 10^{-6}$ in our experiments. The overall architecture for the DL-GAN model is shown in Figure \ref{fig:dlgan-arch} below,

\begin{figure}[hbt!]
    \centering
    \includegraphics[width=\linewidth]{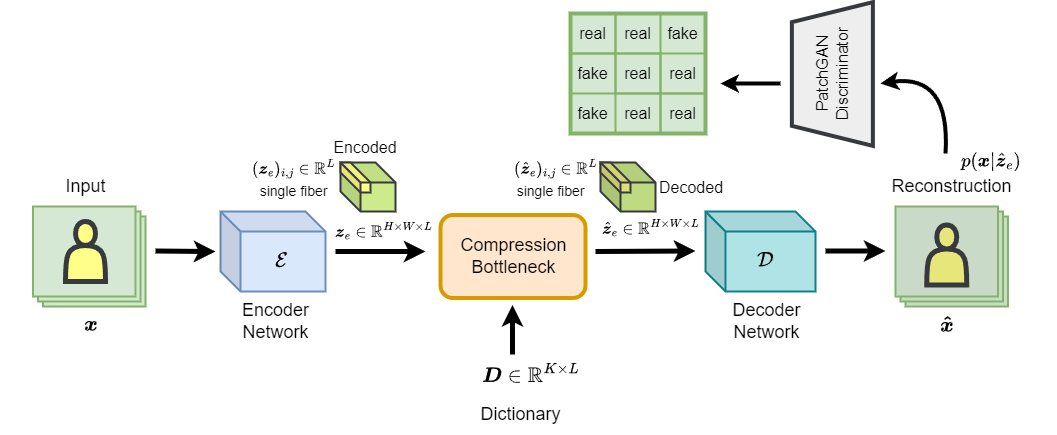}
    \caption{High-level architectural overview of DL-GAN.}
    \label{fig:dlgan-arch}
\end{figure}

\section{Evaluation Methods}

Here to evaluate the reconstruction quality for the images, we adopt the conventional image evaluation metric Peak Signal-to-Noise Ratio (PSNR)~\cite{psnr} and the deep learning based metrics LPIPS~\cite{lpips} loss and the FLIP loss~\cite{flip}, which tends to focus more on perceptual quality instead of pixel by pixel recovery by comparing the image features extracted by deep neural networks instead of the image pixels directly. To evaluate the model's performance on addressing codebook collapse issue as introduced by VQ-family models, we compute the perplexity to measure the usage of either codebook or dictionary atom usage~\cite{CoverThomas}, which is defined as the exponent of average log likelihood of all embedding vectors in an input sequence,
\begin{equation}
    \mathtt{ppl}(\boldsymbol{W}) = \exp \left\{-\frac{1}{N} \sum_{i=1}^{N} \log p(\boldsymbol{w}_i | \boldsymbol{w}_{< i}) \right\},
\end{equation}
where $\mathtt{ppl}(\boldsymbol{W})$ denotes the perplexity of codebook matrix $\boldsymbol{W}$ containing $N$ codewords $\boldsymbol{w}_i, i = 1, 2, \cdots, N$. We use UMAP~\cite{umap} to visualize the vectors.

\section{Experiments}

In our experiments, we adopt a similar lightweight encoder-decoder architecture for our model as proposed in~\cite{vqvae}, with the encoder network consists of $2$ strided convolutional layers with stride $2$ and kernel size $4 \times 4$, followed by two residual blocks of size $3 \times 3$ (implemented as ReLU + $3 \times 3$ convolutional layer + $1 \times 1$ convolutional layer), all having $128$ hidden units. Similarly for the decoder network we have two $3 \times 3$ residual blocks, followed by two transposed convolutions with stride $2$ and kernel size $4 \times 4$. We use the Adam optimizer~\cite{adam} with a learning rate of $1 \times 10^{-4}$ for both VQ-family models and the DL-family models training. We also implement both the codebook in the VQ-family models and the dictionary in the DL-family models as an embedding layer, with dimension of $512 \times 16$, containing $512$ latent embedding vectors of feature dimension $16$. For the VQ-family model training, we adopt a procedure based on the Exponential Moving Average (EMA) with a decay rate of $0.99$, which proves to converge faster and handle codebook collapse better~\cite{vqvae, ema, edvae}. As for the DL-family model training, we use a sparsity level of $5$, which allows $5$ dictionary atoms to be used in the linear combination to represent each latent feature vector.

\begin{figure}[hbt!]
    \centering
    \includegraphics[width=\columnwidth]{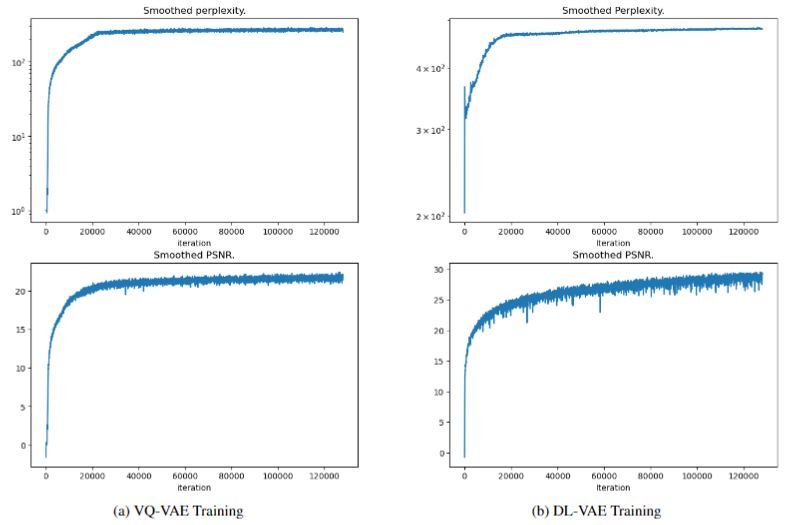}
    \caption{(a) The training evolution of the VQ-VAE model; Figure (b) The training evolution of the DL-VAE model. For both models we evaluate the codebook/dictionary perplexity and the reconstruction PSNR, both in a smoothed fashion using the Savitzky–Golay filter~\cite{savgol}.}
    \label{fig:training}
\end{figure}

For the training experiments, we tested the model VQ-VAE and DL-VAE with the Oxford Flowers~\cite{oxford-flowers} dataset, which contains over $1,000$ high quality flower images, we augment the dataset by extracting random crops of $256 \times 256$ images and train both models for $1,000$ epochs, the results of the training experiments are shown in Figure \ref{fig:training}.

From the training experiments we can see that the DL-VAE converges at a much higher reconstruction PSNR as compared to the VQ-VAE, also the DL-VAE does not suffer codebook collapse, reaching a much higher perplexity value. In particular, the VQ-VAE has a perplexity of 275 with a PSNR of 21.9 whereas our DL-VAE has a perplexity of 507 with a PSNR of 29.1.



\subsection{Latent Space Reconstructions}

Since both models essentially perform latent space reconstruction with different forms of lossy compression, we can evaluate the latent space reconstruction quality from the two compression bottlenecks, here we take results from earlier stages of training ($10$ epochs) so that we can spot on the difference between the two compression bottlenecks on the latent space more easily (when both models converge the differences are not easily visible). Since the latent feature dimension is $16$ which is too high for a sensible visualization, we extract the top singular components from the latent spaces and project them on the grayscale space,

\begin{figure}[hbt!]
    \centering
    \includegraphics[width=0.75\columnwidth]{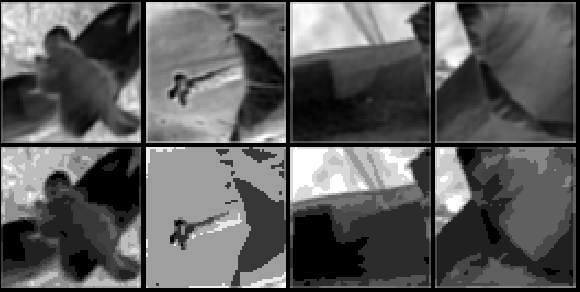}
    \caption{Figures on the first row shows the top singular component from the VQ-VAE encoder output; Figures on the bottom row shows the top singular component from the early stage latent space reconstruction via the Vector Quantization bottleneck.}
    \label{fig:vqvae-latent}
\end{figure}

\begin{figure}[hbt!]
    \centering
    \includegraphics[width=0.75\columnwidth]{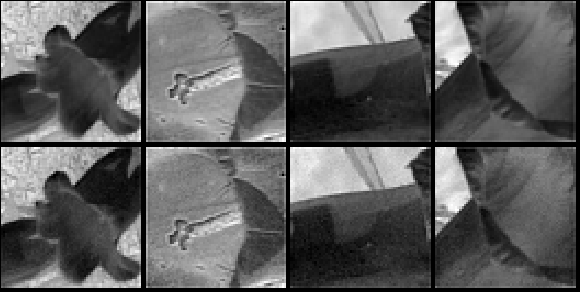}
    \caption{Figures on the first row shows the top singular component from the VQ-VAE encoder output; Figures on the bottom row shows the top singular component from the early stage latent space reconstruction via the Vector Quantization bottleneck.}
    \label{fig:dlvae-latent}
\end{figure}

Note that here from Figure \ref{fig:vqvae-latent} we can clearly see the effect of vector quantization at the early stage of training, with a lot of quantization artifacts in the latent space, however as we can see the reconstructions from the dictionary learning bottleneck are a lot smoother, aligning with our intuition of a more relaxed structural constraint. More concretely, we calculated the reconstruction MSE loss for both models, with the loss of the VQ-VAE model being $0.0701$ and $0.0039$ for the DL-VAE model.

\subsection{The Embedding Visualizations}

\begin{figure}[hbt!]
    \centering
    \includegraphics[width=\columnwidth]{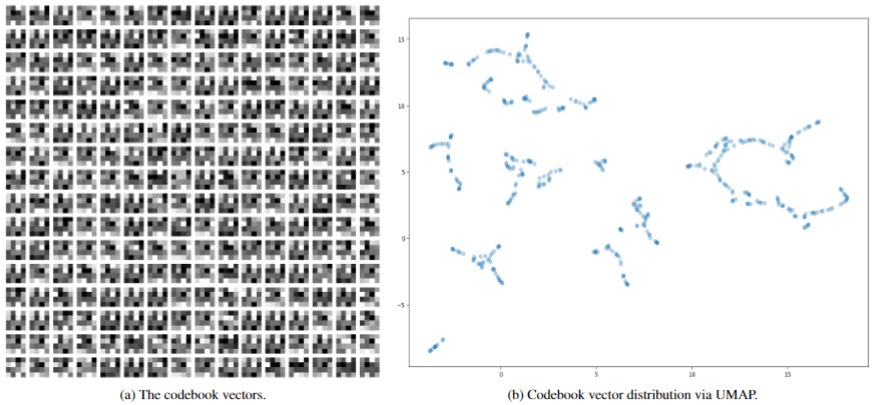}
    \caption{(a) grayscale images of the reshaped codebook vectors. (b) Codebook vector distribution in the embedding space.}
    \label{fig:vq-viz}
\end{figure}

\begin{figure}[hbt!]
    \centering
    \includegraphics[width=\columnwidth]{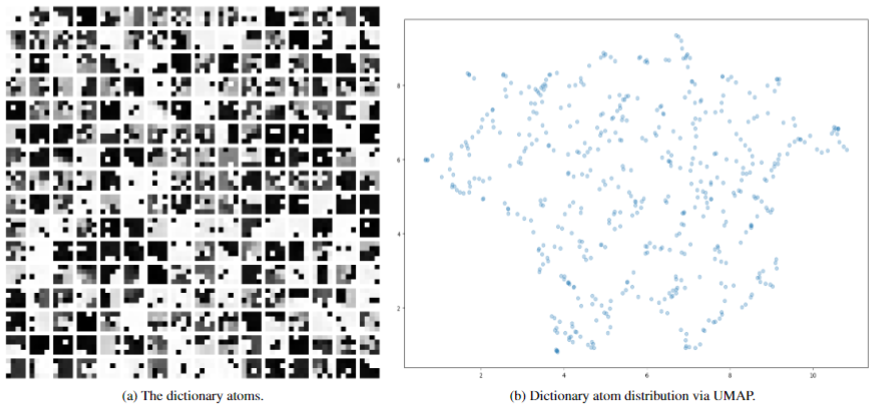}
    \caption{(a) grayscale images of the reshaped dictionary atoms. (b) Dictionary atom distribution in the embedding space.}
    \label{fig:dl-viz}
\end{figure}

To further validate the DL bottleneck's ability in circumventing codebook collapse, we reshape the $16$ dimensional embedding vectors into $4 \time 4$ patches and visualize them in grayscale images, we have also provided a visualization of the embedding vector distribution in the latent space using UMAP~\cite{umap}. The visualizations regarding the VQ-VAE and the DL-VAE models are shown in Figure \ref{fig:vq-viz} and Figure \ref{fig:dl-viz} respectively.

From the visualizations, we can clearly see the codebook collapse phenomenon in the VQ-VAE model, where all the codebook vectors tend to cluster to each other and are rather similar in the look, while the dictionary atoms are distributed in a more spread out fashion with much different looks from each other.

\subsection{The Reconstructions}

Here are the reconstructions from our trained models on Oxford Flowers dataset~\cite{oxford-flowers}, to highlight the differences between the reconstructions and the original images, we apply the FLIP heatmap to both model outputs, where the darker the region in the heatmap, the more fidelity in reconstruction, and the discrepancies in the reconstructions are thus visualized in highlighted regions.

\begin{figure}[hbt!]
    \centering
    \includegraphics[width=\columnwidth]{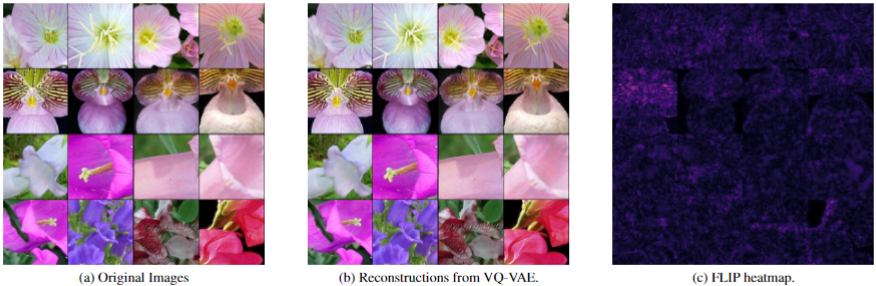}
    \caption{(a) Original Images. (b) Reconstructions from VQ-VAE. (c) FLIP heatmap.}
    \label{fig:vq-results}
\end{figure}

\begin{figure}[hbt!]
    \centering
    \includegraphics[width=\columnwidth]{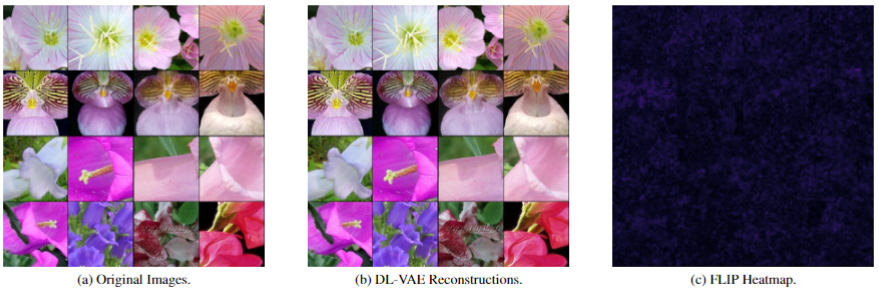}
    \caption{(a) Original Images. (b) Reconstructions from DL-VAE. (c) FLIP heatmap.}
    \label{fig:dl-results}
\end{figure}

\begin{figure}[hbt!]
    \centering
    \includegraphics[width=0.75\columnwidth]{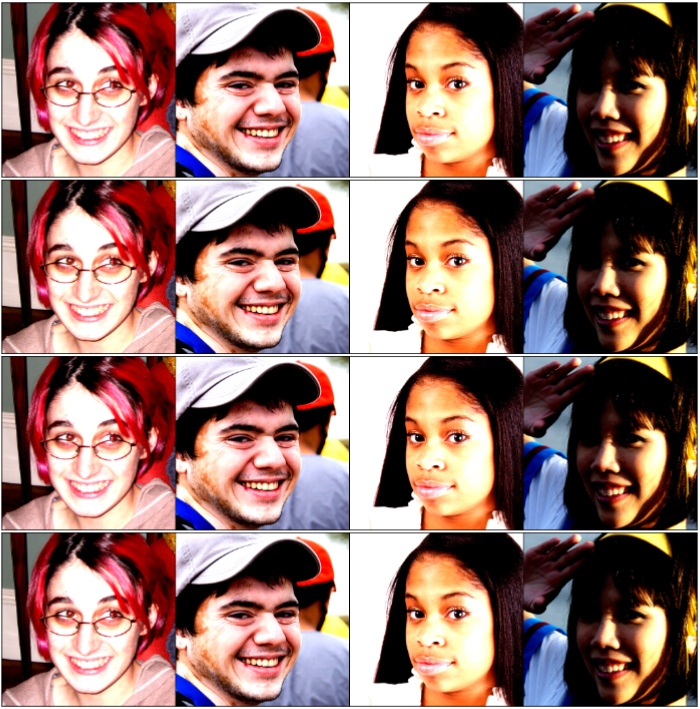}
    \caption{Reconstruction visualizations of VQ-GAN, DL-VAE, and DL-GAN with sparsity level $5$, the first row are original images from the FFHQ dataset, while the second row are the VQ-GAN reconstructions, the third row are the DL-VAE reconstructions, and the fourth row are the DL-GAN reconstructions.}
    \label{fig:ffhq-results}
\end{figure}

\begin{figure}[hbt!]
    \centering
    \includegraphics[width=0.75\linewidth]{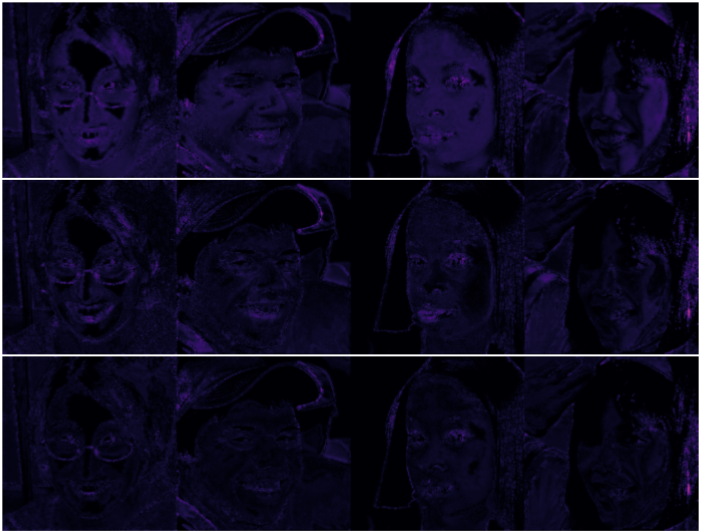}
    \caption{FLIP heatmap visualizations of VQ-GAN, DL-VAE, and DL-GAN with sparsity level $5$, the images in the first row are the FLIP heatmap between the original images and the VQ-GAN reconstructions, the images in the second row are the FLIP heatmap between the original images and the DL-VAE reconstructions, the images in the third row are the FLIP heatmap between the original images and the DL-GAN reconstructions.}
    \label{fig:ffhq-flip}
\end{figure}

Some further reconstructions results on the FFHQ dataset~\cite{ffhq}, along with the VQ-GAN and the proposed DL-GAN models are shown in the Figure \ref{fig:ffhq-results} and the Figure \ref{fig:ffhq-flip},

Here is the table summarizing the model reconstruction PSNR on various datasets,
\begin{table}
  \centering
  {\small{
  \begin{tabular}{@{}lc@{}lc@{}@{}lc@{}lc@{}}
    \toprule
    Model & MNIST &\quad CIFAR10 &Oxford Flowers &\quad FFHQ\\
    \midrule
    VQ-VAE & $18.1$ &\qquad $22.9$ &$21.9$ &\quad $23.8$\\
    VQ-GAN & $19.3$ &\qquad $23.2$ &$22.5$ &\quad $24.6$\\
    DL-VAE & $25.9$ &\qquad $29.2$ &$29.1$ &\quad $32.2$\\
    DL-GAN & $\textbf{27.2}$ &\qquad $\textbf{31.1}$ &$\textbf{31.4}$ &\quad $\textbf{33.7}$\\
    \bottomrule
  \end{tabular}
  }}
  \caption{Reconstruction PSNRs attained from training on various datasets. All PSNR values are reported upon training convergence.}
  \label{tab:train-psnrs}
\end{table}

\section{Conclusion and Future Work}

In this paper, we examined the effectiveness of latent space representation learning by sparse dictionary learning instead of the vector quantization, and demonstrated that the effectiveness of vector quantization in previous models might not be the only answer to effective latent representation learning. The derived models, DL-VAE and DL-GAN have demonstrated solid performance in terms of latent space representation power and robustness to codebook collapse. Also be aware that the sparsity assumtion we are using, though demonstrating strong performance against the VQ-family models, still is not the only best way to represent the latent space, this idea leaves huge room for further research in terms of the representation learning in latent space and using this learned representation to support other state-of-the-art generative models ~\cite{vqgan, ldm, gpt}, etc.



\clearpage

{\small
\bibliographystyle{ieee_fullname}
\bibliography{main}
}

\medskip

\clearpage

\title{LASERS: LAtent Space Encoding for Representations with Sparsity for Generative Modeling Supplementary Material}

\maketitle

\section{The Sparse Coding Algorithm}

In our implementations \footnote{Please feel free to check our \href{https://anonymous.4open.science/r/dlgan-0CCD}{GitHub Repository} for our codebase and experiments setups.} of the DL-family models (DL-VAE and DL-GAN), we adopt the Batch-OMP \cite{efficient-ksvd} procedure as our sparse coding algorithm, which has demonstrated superior performance in terms of efficiency and convergence as compared to other greedy pursuits and thresholding based methods \cite{omp, cs, ista, fista, lista}. Specifically, given the dictionary $\boldsymbol{D}$, the support set $\mathcal{T}$ ($|\mathcal{T}| = S$ for sparsity level $S$), and the residual $\boldsymbol{r}$ the classic greedy OMP algorithm usually involves the computation \cite{cs, efficient-ksvd},
\begin{align*}
    \hat{\boldsymbol{\gamma}}_{\mathcal{T}} &= \boldsymbol{D}_{\mathcal{T}}^\dagger \boldsymbol{r} \\
    &= (\boldsymbol{D}_{\mathcal{T}}^\top \boldsymbol{D}_{\mathcal{T}})^{-1} \boldsymbol{D}_{\mathcal{T}}^\top \boldsymbol{r},
\end{align*}
requiring the expensive matrix inversion of $\boldsymbol{D}_{\mathcal{T}}^\top \boldsymbol{D}_{\mathcal{T}}$. However, note that $\boldsymbol{D}_{\mathcal{T}}^\top \boldsymbol{D}_{\mathcal{T}}$ is actually a symmetric positive definite matrix and thus can be addressed more efficiently by progressive Cholesky factorization \cite{matrix-computations, efficient-ksvd}. Also the atom selection step at each iteration does not require knowing $\boldsymbol{r}$ or $\boldsymbol{\gamma}$ explicitly, but rather only $\boldsymbol{D}^\top \boldsymbol{r}$. Hence we can seek to replace the explicit computation of $\boldsymbol{r}$ and its multiplication by $\boldsymbol{D}^\top$ with a lower-cost computation $\boldsymbol{D}^\top \boldsymbol{r}$. To achieve that, denote $\boldsymbol{\alpha} = \boldsymbol{D}^\top \boldsymbol{r}$, $\boldsymbol{\alpha}^0 = \boldsymbol{D}^\top \boldsymbol{y}$ with $\boldsymbol{y}$ denoting the initial signal input, and $\boldsymbol{G} = \boldsymbol{D}^\top \boldsymbol{D}$, we then have,
\begin{align}
    \boldsymbol{\alpha} &= \boldsymbol{D}^\top (\boldsymbol{y} - \boldsymbol{D}_{\mathcal{T}}\boldsymbol{D}_{\mathcal{T}}^\dagger \boldsymbol{y}) \nonumber\\
    &= \boldsymbol{\alpha}^0 - \boldsymbol{G}_{\mathcal{T}}\boldsymbol{D}_{\mathcal{T}}^\dagger \boldsymbol{y} \nonumber\\
    &= \boldsymbol{\alpha}^0 - \boldsymbol{G}_{\mathcal{T}} (\boldsymbol{D}_{\mathcal{T}}^\top \boldsymbol{D}_{\mathcal{T}})^{-1}\boldsymbol{D}_{\mathcal{T}}^\top \boldsymbol{y}.
\end{align}

The algorithm pseudocode is shown below \cite{efficient-ksvd},
\begin{algorithm}
\caption{Batch Orthogonal Matching Pursuit (Batch-OMP)~\cite{cs}}\label{alg:batch-omp}
\KwIn{one column of the flattened encoder output $(\Bar{\boldsymbol{z}}_e)_n, n \in [N]$ where $N = H \times W$, embedding/dictionary matrix $\boldsymbol{D}$, and sparsity level $S$}
\Init{$\boldsymbol{\alpha}^0 = \boldsymbol{D}^\top(\Bar{\boldsymbol{z}}_e)_n$, initial gram matrix $\boldsymbol{G}^{(0)} = \boldsymbol{D}^\top \boldsymbol{D}$}{}
\For{$s \gets 1; s \gets s + 1; s \gets S$}{
    Calculate the following,
    \begin{equation}
        j^{(s)} \gets \argmax_{j} | \boldsymbol{\alpha}_j^{(s)} |
    \end{equation}
    which essentially selects elements based on the magnitude of the elements of $\boldsymbol{\alpha}^{(s)}$\;
    \If{$s > 1$}{
    Solve for $\boldsymbol{\omega}^{(s)}$ as the solution of linear equation $\boldsymbol{L}^{(s - 1)}\boldsymbol{\omega}^{(s)} = \boldsymbol{G}^{(s - 1)}_{\tau, \hat{j}}$, and then construct
    \begin{equation*}
        \boldsymbol{L}^{(s)} \gets \begin{bmatrix}
            \boldsymbol{L}^{(s - 1)} &\boldsymbol{0}\\
            (\boldsymbol{\omega}^{(s)})^\top &\sqrt{1 - (\boldsymbol{\omega}^{(s)})^\top \boldsymbol{\omega}^{(s)}}
        \end{bmatrix}
    \end{equation*}
    }
    Append the support set,
    \begin{equation}
        \tau^{(s)} \gets \tau^{(s - 1)} \cup j^{(s)}
    \end{equation}
    which updates the support set corresponding to the signal estimate\;
    Update the signal estimate by solving the linear system,
    \begin{align}
        \hat{\boldsymbol{\gamma}}^{(s)} &\gets \text{ Solve for } \boldsymbol{L}^{(s)}(\boldsymbol{L}^{(s)})^\top \boldsymbol{\gamma} = \boldsymbol{\alpha}^0_{\tau^{(s)}}\nonumber\\
        &\quad \textit{subject to}\quad \mathrm{supp}\{ \boldsymbol{\gamma} \} = \tau^{(s)}
    \end{align}
    note that here the linear solver is performed over all $\boldsymbol{x}$ with support $\tau^{(s)}$\;
    Calculate
    \begin{equation}
        \boldsymbol{\alpha}^{(s)} \gets \boldsymbol{\alpha}^0 - \boldsymbol{G}_{\tau^{(s)}} \hat{\boldsymbol{\gamma}}^{(s)}_{\tau^{(s)}}
    \end{equation}
    to update the product term $\boldsymbol{\alpha}$\;
}
\KwOut{sparse code $\hat{\boldsymbol{\gamma}}^{(S)}$.}
\end{algorithm}

\section{Online Dictionary Learning Based on Block-Coordinate Descent}

Apart from the implicit dictionary learning method mentioned in the paper, we can also break the dictionary learning process into two stages, with the first stage being the sparse coding stage and the second stage being a learning rate free online dictionary update via Block-Coordinate Descent \cite{online-dl}. Specifically, at each training iteration, after passing the input from the encoder network, we first assume that the dictionary is fixed and perform the sparse coding stage (Algorithm \ref{alg:batch-omp}) to compute the sparse code associated with the training sample. Then we adopt a Block-Coordinate Descent procedure to update the dictionary atoms based on the calculated sparse code. Specifically, on training iteration $t$, consider the sparse code $\boldsymbol{\gamma}^{(t)}$ computed from the sparse coding stage and the minibatch input encodings $\boldsymbol{z}_e^{(t)} \in \reals^{B \times H \times W \times L}$ from the encoder, we first flatten the encodings into $\Bar{\boldsymbol{z}}_e^{(t)} \in \reals^{L \times BHW}$ shape, denote $\xi = BHW$, we then calculate the moving average of the matrices from gradient computations,
\begin{align}
    \boldsymbol{A}^{(t)} &= \beta \boldsymbol{A}^{(t - 1)} + \boldsymbol{\gamma}^{(t)} \left(\boldsymbol{\gamma}^{(t)}\right)^\top,\\
    \boldsymbol{B}^{(t)} &= \beta \boldsymbol{B}^{(t - 1)} + \boldsymbol{z}_e^{(t)} \left(\boldsymbol{\gamma}^{(t)}\right)^\top,
\end{align}
where $\beta$ is an adaptive parameter for better convergence in practice \cite{online-dl},
\begin{equation}
    \beta = \frac{\theta + 1 - \xi}{\theta + 1},
\end{equation}
where the parameter $\theta$ is computed as follows,
\begin{equation}
    \theta = \begin{cases}
        t \xi, &\text{if } t < \xi,\\
        \xi^2 + t - \xi, &\text{if } t \geq \xi.
    \end{cases}
\end{equation}
Then we perform one iteration of the Block-Coordinate Descent algorithm on all the codebook embedding columns, which is sufficiently accurate for the online setting \cite{online-dl},
\begin{align}
    \boldsymbol{D}_{:, j} &\gets \frac{1}{\boldsymbol{A}_{jj}} \left(\boldsymbol{B}_{:, j} - \boldsymbol{D}\boldsymbol{A}_{:, j} + \boldsymbol{D}_{:, j}\boldsymbol{A}_{jj} \right)\\
    \boldsymbol{D}_{:, j} &\gets \boldsymbol{D}_{:, j} / \| \boldsymbol{D}_{:, j} \|_2.
\end{align}
With sufficient number of training iterations and an accurate enough sparse coding stage, we expect the model training to converge. However, in practice this algorithm is slow when the number of dictionary atoms is large and does not yield necessarily superior performance compared to our implicit method, thus we do not adopt this method in the actual implementation but still we provide the implementation in the codebase for future potential improvements.

\medskip

\section{More Results from Training Experiments}

Here we provide some additional results from our training experiments. Here in the following sections we majorly focus on comparing the VQ and DL family models to evaluate the performance of the compression bottleneck, hence models like VQ-VAE2 \cite{vqvae2} that stacks two bottlenecks together are not relevant in comparing one unit of the bottleneck performance, also models like ViT-VQGAN \cite{improved-vqgan} that use different encoder and decoder architectures while keeping the compression bottleneck intact are also not relevant for our comparisons. We still use the same hyperparameter setting for both of the models. in the main paper, with codebook/dictionary of size $512 \times 16$ and sparsity level $5$ for the DL family models, the encoder and decoder architectures for both of the VQ and DL models are also the same as mentioned in the main paper, we also implemented the same PatchGAN discriminator architecture as suggested in \cite{vqgan, patchgan}. For the CIFAR10 dataset \cite{cifar10}, we performed a $50,000 / 10,000$ train-test split and trained with a batch size of $256$ for $200$ epochs, with additional $20$ epochs when there's discriminator training, the results are as follows,

\begin{figure}
    \centering
    \includegraphics[width=0.75\columnwidth]{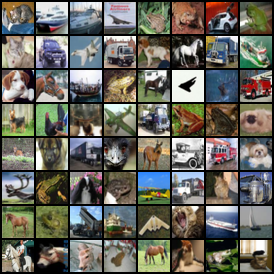}
    \caption{Target Images.}
    \label{fig:cifar10-target}
\end{figure}

\begin{figure}
    \centering
    \includegraphics[width=0.75\columnwidth]{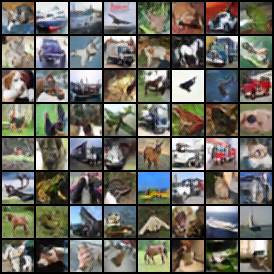}
    \caption{VQ-GAN Reconstructions.}
    \label{fig:vqgan-cifar10}
\end{figure}

\begin{figure}
    \centering
    \includegraphics[width=0.75\linewidth]{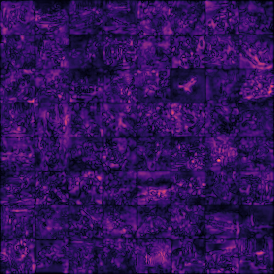}
    \caption{FLIP Heatmap of the VQ-GAN Reconstructions.}
    \label{fig:vqgan-cifar10-flip}
\end{figure}

\begin{figure}
    \centering
    \includegraphics[width=0.75\columnwidth]{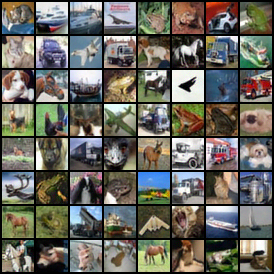}
    \caption{DL-GAN Reconstructions.}
    \label{fig:dlgan-cifar10}
\end{figure}

\begin{figure}
    \centering
    \includegraphics[width=0.75\linewidth]{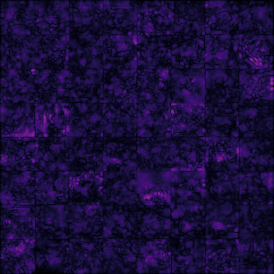}
    \caption{FLIP Heatmap of the DL-GAN Reconstructions.}
    \label{fig:dlgan-cifar10-flip}
\end{figure}

Table \ref{tab:train-cifar10} summarizes the numeric evaluation results of the two models on CIFAR10 dataset,

\begin{table}
  \centering
  {\small{
  \begin{tabular}{@{}lc@{}lc@{}@{}lc@{}lc@{}}
    \toprule
    Model & PSNR &\qquad LPIPS &\quad FLIP\\
    \midrule
    VQ-VAE & $22.36$ &\qquad $0.1629$ &\quad $0.05842$\\
    VQ-VAE & $26.47$ &\qquad $0.07492$ &\quad $0.05358$\\
    DL-VAE & $28.81$ &\qquad $0.02903$ &\quad $0.05181$\\
    DL-VAE & $\textbf{30.01}$ &\qquad $\textbf{0.01745}$ &\quad $\textbf{0.04792}$\\
    \bottomrule
  \end{tabular}
  }}
  \caption{Reconstruction evaluations on the CIFAR10 dataset.}
  \label{tab:train-cifar10}
\end{table}

We have also evaluated our models on the Flickr-Faces HQ (FFHQ) dataset \cite{ffhq}, which contains $70,000$ high quality $1024 \times 1024$ RGB images of human faces. For the training setup, we performed a $60,000/10,000$ train-test split and downsampled the images to $512 \times 512$ to fit in our GPU memory. We trained both VQ and DL family models on the dataset with a batch size of $8$ for $20$ epochs, with additional $5$ epochs for discriminator training. The results are shown in the figures below,

\begin{figure*}
    \centering
    \includegraphics[width=\textwidth]{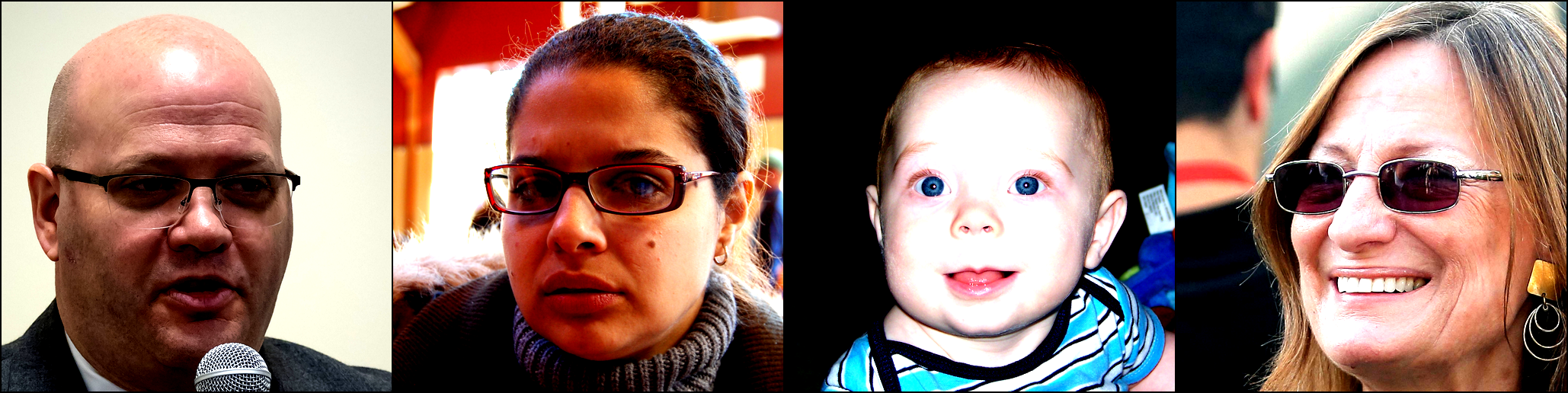}
    \caption{Target Images.}
    \label{fig:ffhq-target}
\end{figure*}

\begin{figure*}
    \centering
    \includegraphics[width=\textwidth]{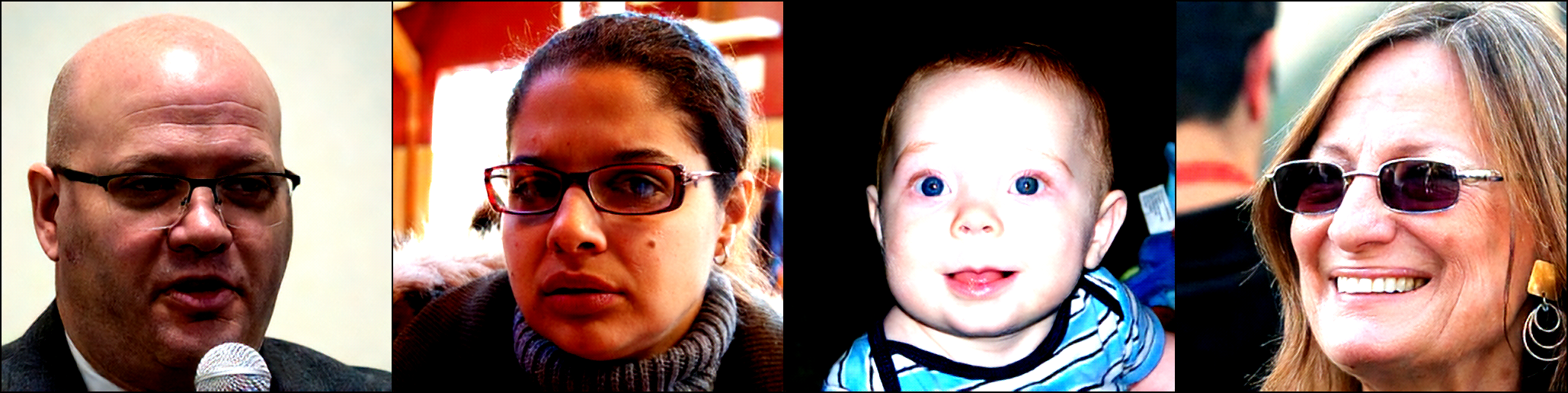}
    \caption{VQ-GAN Reconstructions.}
    \label{fig:vqgan-ffhq}
\end{figure*}

\begin{figure*}
    \centering
    \includegraphics[width=\textwidth]{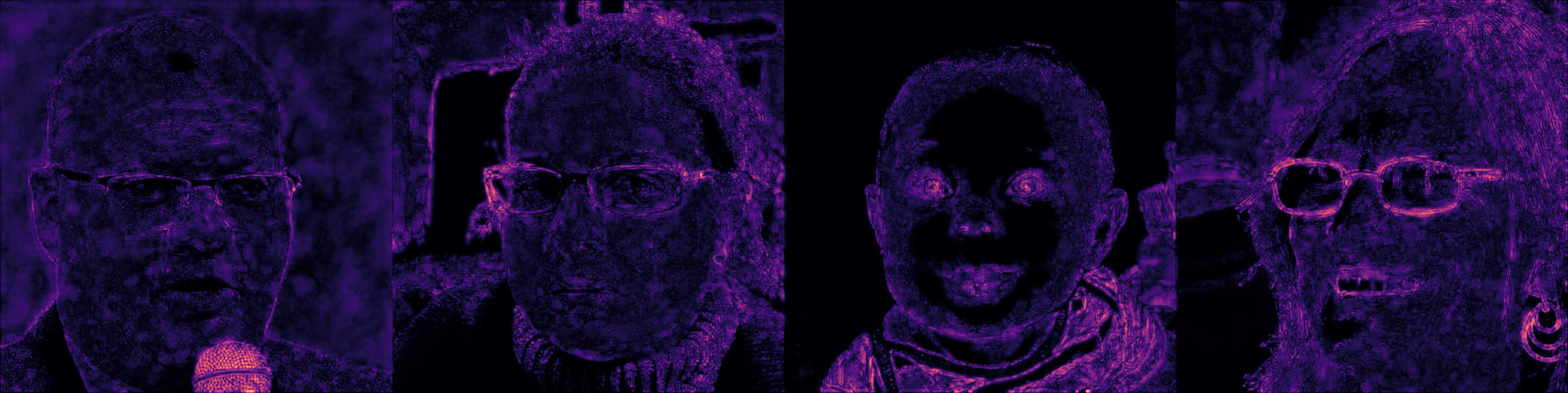}
    \caption{FLIP Heatmap of the VQ-GAN Reconstructions.}
    \label{fig:vqgan-ffhq-flip}
\end{figure*}

\begin{figure*}
    \centering
    \includegraphics[width=\textwidth]{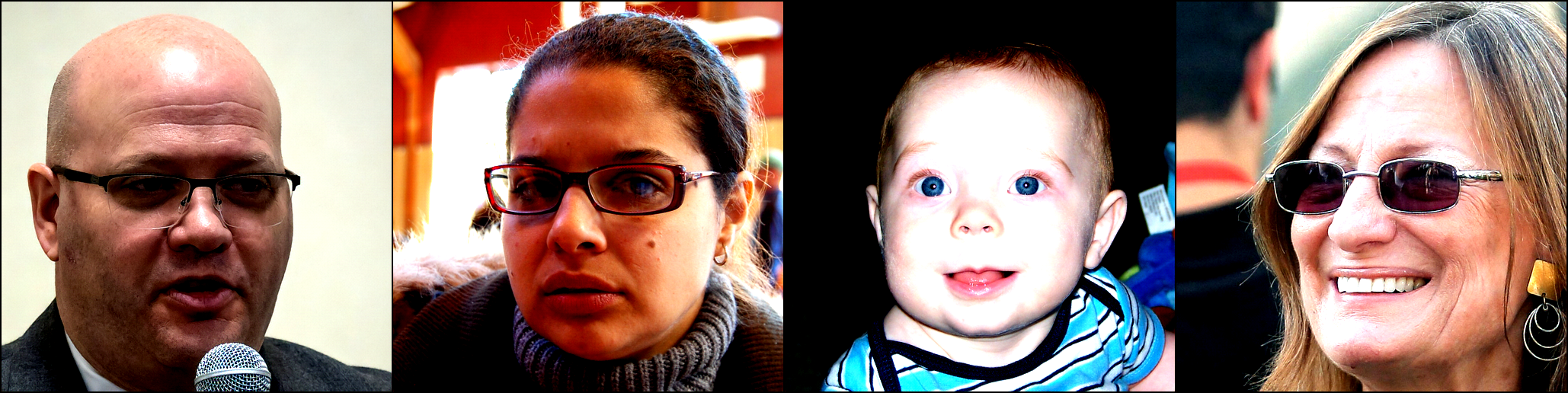}
    \caption{DL-GAN Reconstructions.}
    \label{fig:dlgan-ffhq}
\end{figure*}

\begin{figure*}
    \centering
    \includegraphics[width=\textwidth]{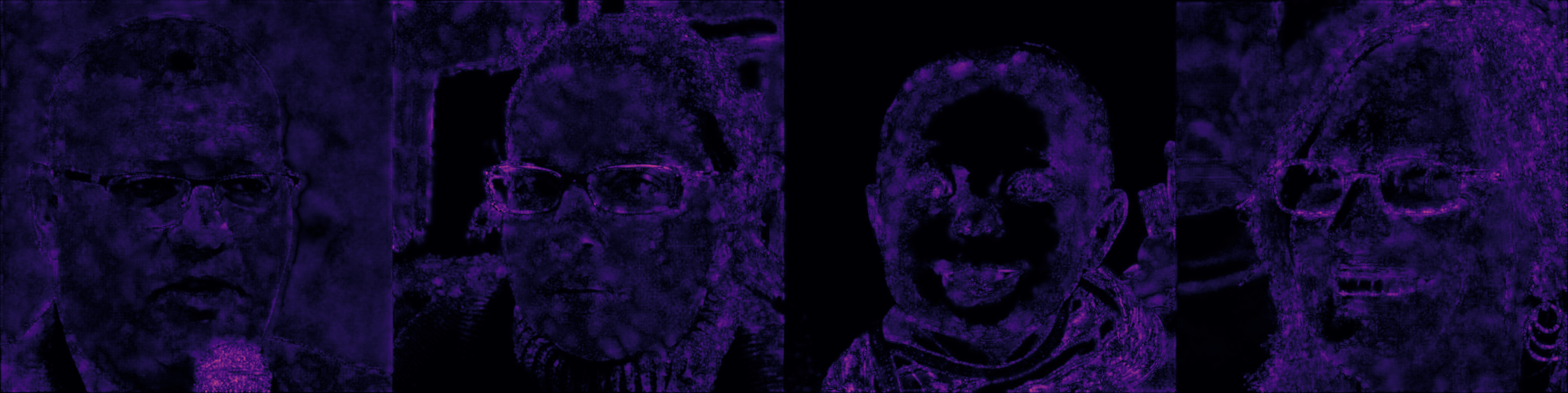}
    \caption{FLIP Heatmap of the DL-GAN Reconstructions.}
    \label{fig:dlgan-ffhq-flip}
\end{figure*}

Table \ref{tab:train-ffhq} summarizes the numeric evaluation results of the two models on FFHQ dataset

\begin{table}
  \centering
  {\small{
  \begin{tabular}{@{}lc@{}lc@{}@{}lc@{}lc@{}}
    \toprule
    Model & PSNR &\qquad LPIPS &\quad FLIP\\
    \midrule
    VQ-VAE & $23.45$ &\qquad $0.1576$ &\quad $0.09929$\\
    VQ-GAN & $25.67$ &\qquad $0.1029$ &\quad $0.08244$\\
    DL-VAE & $30.80$ &\qquad $0.03749$ &\quad $0.07783$\\
    DL-GAN & $\textbf{32.93}$ &\qquad $\textbf{0.02507}$ &\quad $\textbf{0.06427}$\\
    \bottomrule
  \end{tabular}
  }}
  \caption{Reconstruction evaluations on the CIFAR10 dataset.}
  \label{tab:train-ffhq}
\end{table}

From the above training experiment results, we can see the DL family models demonstrate superior performance compared to VQ family models by a comfortable margin.

\medskip

\section{Impact of Codebook/Dictionary Sizes}
\label{sec:cbd}

To analyze the impact of codebook/dictionary sizes on the models, we evaluated both VQ and DL family models on the CIFAR10 dataset \cite{cifar10} with varying codebook/dictionary sizes for $100$ epochs, with additional $20$ epochs for discriminator training, the resulting reconstruction PSNRs are summarized in the table \ref{tab:cbd}. Note that for all the DL family models experiments we use a sparsity level of $4$.

\begin{table}
  \centering
  {\small{
  \begin{tabular}{@{}lc@{}lc@{}@{}lc@{}lc@{}}
    \toprule
    Size & VQ-VAE &\quad VQ-GAN &DL-VAE &\quad DL-GAN\\
    \midrule
    $64$ & $22.53$ &\qquad $23.69$ &\quad $26.79$ &\qquad $\textbf{27.85}$\\
    $256$ & $23.59$ &\qquad $25.74$ &\quad $28.03$ &\qquad $\textbf{29.91}$\\
    $512$ & $22.36$ &\qquad $24.93$ &\quad $28.95$ &\qquad $\textbf{30.04}$\\
    \bottomrule
  \end{tabular}
  }}
  \caption{Reconstruction evaluations on the CIFAR10 dataset on varying codebook/dictionary sizes.}
  \label{tab:cbd}
\end{table}

\medskip

\section{Ablation Studies on Sparsity Levels}
\label{sec:ablation}

We perform ablation studies with respect to the sparsity level of the Dictionary Learning Compression Bottleneck, we evaluated the DL-VAE model on the CIFAR10 dataset \cite{cifar10} with varying sparsity levels with fixed dictionary size as $512$ for $100$ epochs, the resulting reconstruction evaluations are summarized in the table \ref{tab:abl}.

\begin{table}
  \centering
  {\small{
  \begin{tabular}{@{}lc@{}lc@{}@{}lc@{}lc@{}}
    \toprule
    Sparsity & PSNR &\quad Perplexity\\
    \midrule
    $2$ & $25.63$ &\qquad $504$\\
    $4$ & $28.95$ &\qquad $507$\\
    $5$ & $29.33$ &\qquad $498$\\
    $8$ & $31.32$ &\qquad $455$\\
    $10$ & $31.96$ &\qquad $428$\\
    \bottomrule
  \end{tabular}
  }}
  \caption{Reconstruction evaluations on the CIFAR10 dataset on varying sparsity levels.}
  \label{tab:abl}
\end{table}

From the above evaluations we can see that the DL-VAE model gains more representation power as the sparsity level goes up, but loses dictionary perplexity at the mean time, we contend that including more dictionary atoms for each latent feature vector gives the dictionary atoms a tendency to become more similar to each other.

\medskip

\section{Downstream Generation}
\label{sec:downstream}

In the following sections we evaluate our model with downstream generation tasks such as super-resolution, inpainting, and text-to-image generation with Stable Diffusion \cite{ldm}. Note that the purpose of this section is not trying to prove our model beats the state-of-the-art generative models, but rather to show the versatile generation power gained by simply restructuring the latent space, we'll focus on the VQ-VAE and DL-VAE model training for the following sections to compare their compression bottleneck performance, and we purposefully disallow the discriminator training so that the additional generation power from the discriminator will not interfere our evaluations on the compression bottleneck performance.

\medskip

\subsection{Single Image Super Resolution Experiments}

For the single image super resolution experiments, we evaluated our models on the Oxford Flowers dataset \cite{oxford-flowers} by first downsampling the images to $64 \times 64$ dimension and then upscaling to $256 \times 256$ dimension via bicubic degradation \cite{bicubic} as the input data, with the original images as the target super resolved images. We trained the VQ-VAE and DL-VAE models on the same dataset with a batch size of $8$ for $2,000$ epochs. For evaluation we also use the Fr\'echet Inception Score (FID, the lower the better) to measure the generation quality \cite{fid}. The results are as follows,

\begin{figure*}[hbt!]
    \centering
    \includegraphics[width=0.25\textwidth]{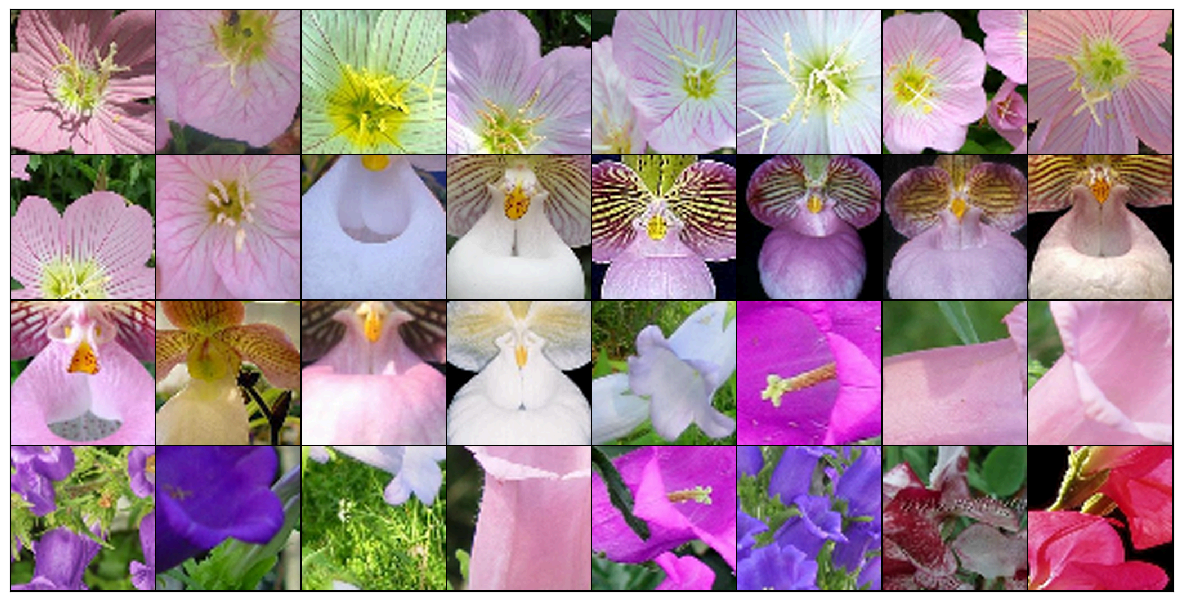}
    \caption{Bicubic Downsampled Low Resolution Inputs.}
    \label{fig:lr-inputs}
\end{figure*}

\begin{figure*}[hbt!]
    \centering
    \includegraphics[width=\textwidth]{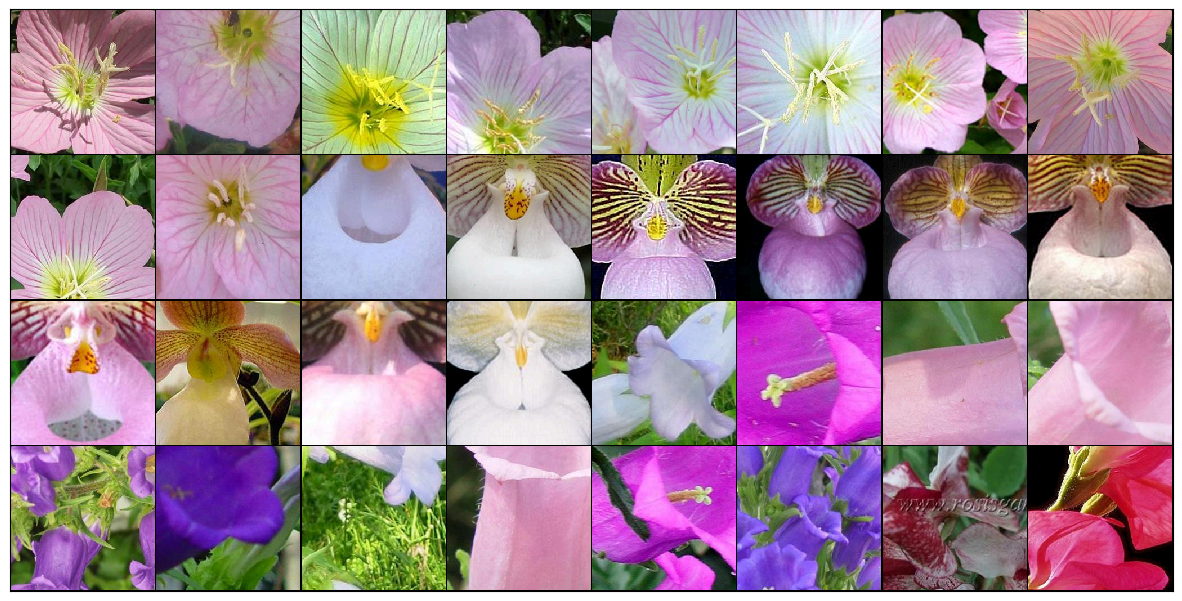}
    \caption{Target Images for Single Image Super Resolution.}
    \label{fig:sr-targets}
\end{figure*}

\clearpage

\begin{figure*}[hbt!]
    \centering
    \includegraphics[width=\textwidth]{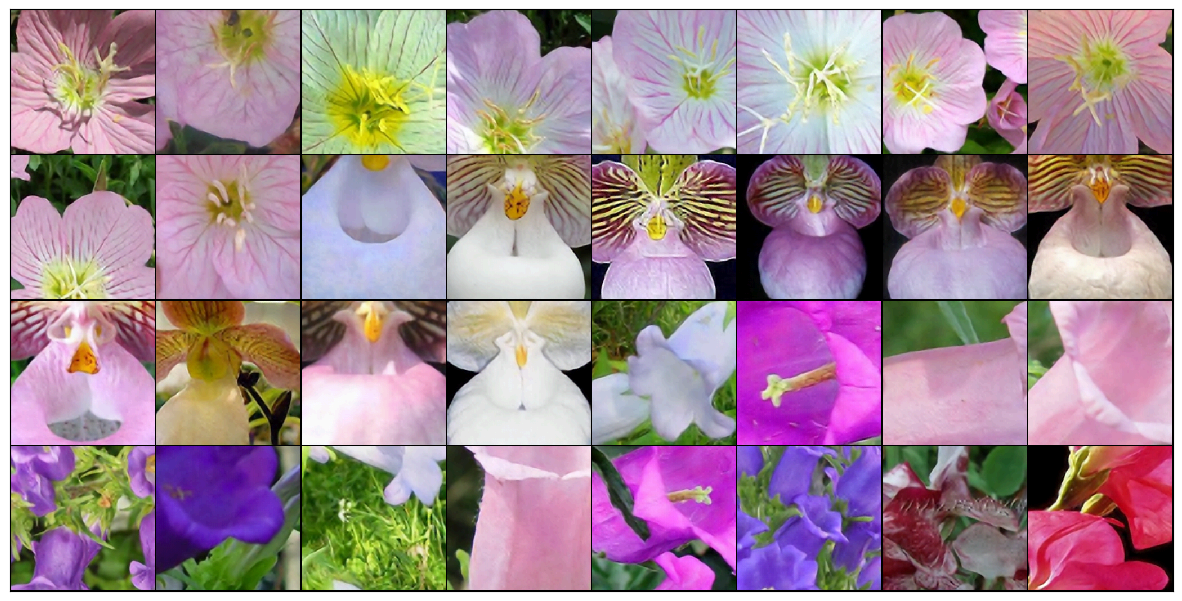}
    \caption{Super Resolved Images from VQ-VAE}
    \label{fig:vqgan-sr}
\end{figure*}

\begin{figure*}
    \centering
    \includegraphics[width=\textwidth]{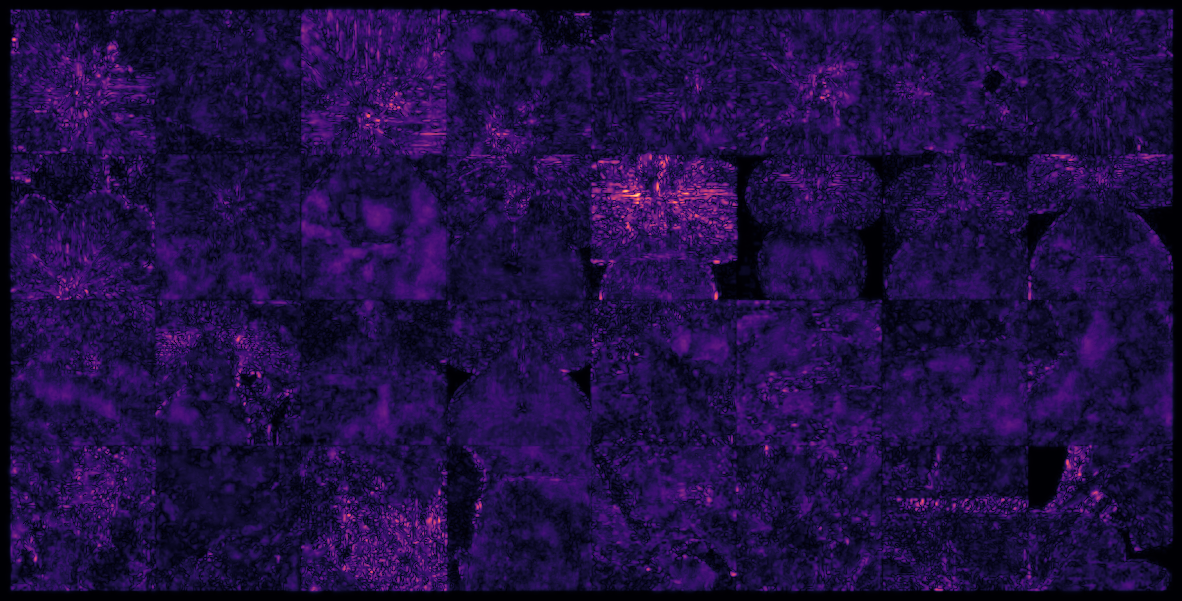}
    \caption{Flip Heatmap Evaluations for the VQ-VAE Super Resolution.}
    \label{fig:vqgan-sr-flip}
\end{figure*}

\begin{figure*}
    \centering
    \includegraphics[width=\textwidth]{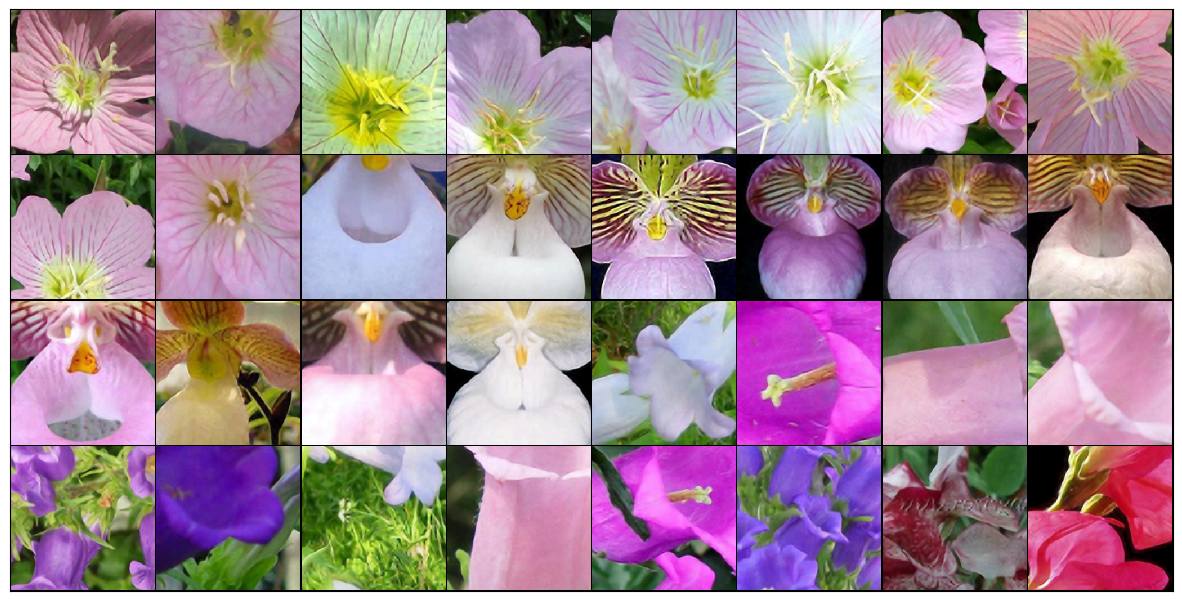}
    \caption{Super Resolved Images from DL-VAE}
    \label{fig:dlgan-sr}
\end{figure*}

\begin{figure*}
    \centering
    \includegraphics[width=\textwidth]{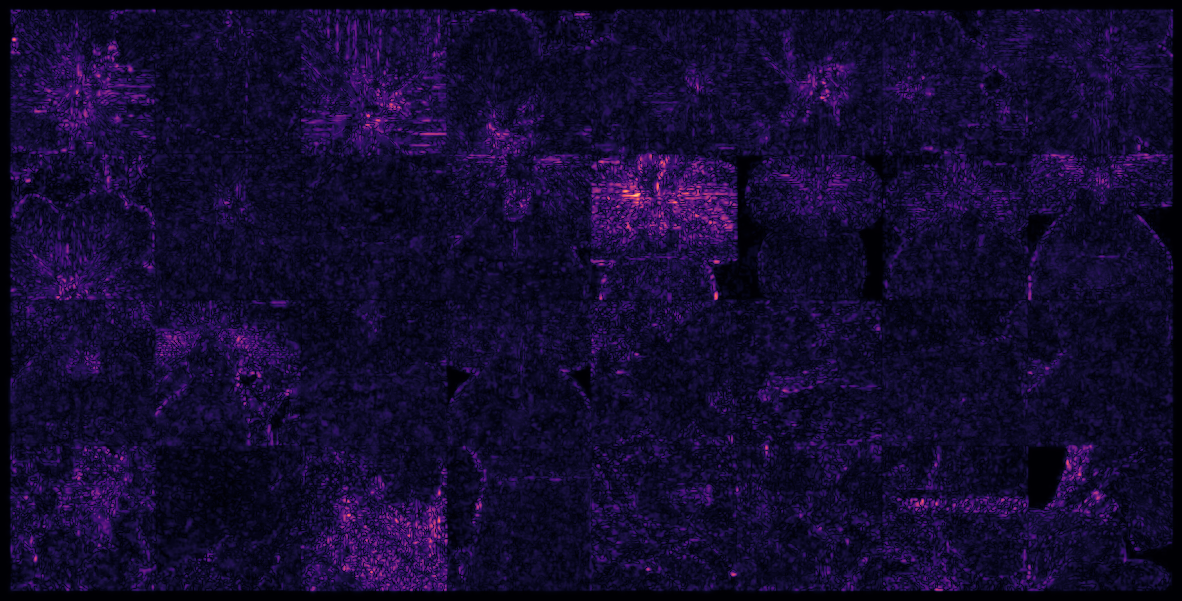}
    \caption{Flip Heatmap Evaluations for the DL-VAE Super Resolution.}
    \label{fig:dlgan-sr-flip}
\end{figure*}

Table \ref{tab:sr} is a brief summary of the quantitative evaluations of the super resolution experiments for both models.

\begin{table}
  \centering
  {\small{
  \begin{tabular}{@{}lc@{}lc@{}@{}lc@{}lc@{}}
    \toprule
    Model & PSNR &\qquad FLIP &\quad FID\\
    \midrule
    VQ-VAE & $28.78$ &\quad\: $0.1219$ &\quad $33.89$\\
    DL-VAE & $\textbf{29.23}$ &\quad\: $\textbf{0.08767}$ &\quad $\textbf{23.31}$\\
    \bottomrule
  \end{tabular}
  }}
  \caption{Single image super resolution on Oxford Flowers dataset.}
  \label{tab:sr}
\end{table}

From the results we can see the DL-VAE model provides more representation power as compared to VQ-VAE in the latent space while all the other model architectural components (encoder, decoder networks) remain the same.

\medskip

\subsection{Inpainting}

To evaluate the latent representation power of the compression bottleneck in terms of inpainting, we trained the VQ-VAE and DL-VAE models on the Oxford Flowers dataset for the same setting as before for $1,000$ epochs. For the input images, we apply a square bitmask that's $0.25$ fraction of the image dimension in the center of the images to mask out the pixels (set the pixel values to $0$). The results are as follows, we have also provided the reconstructed latent space top singular component visualizations to see how the masking affects the latent space,

\begin{figure*}
    \centering
    \includegraphics[width=\textwidth]{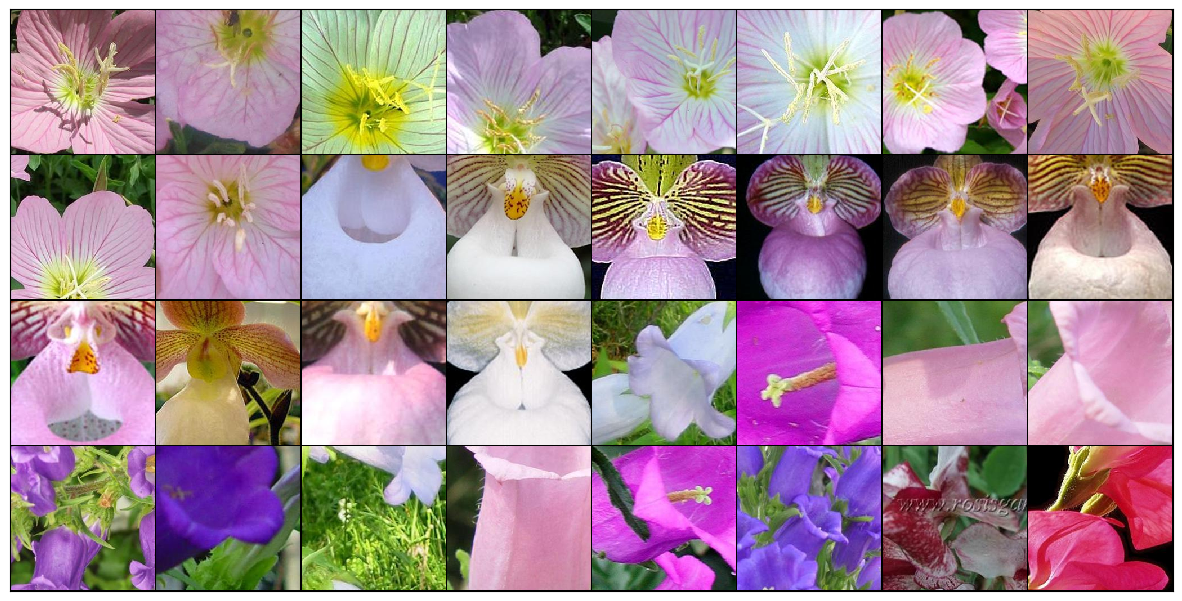}
    \caption{Original Input Images for Inpainting.}
    \label{fig:inpainting-target}
\end{figure*}

\begin{figure*}
    \centering
    \includegraphics[width=\textwidth]{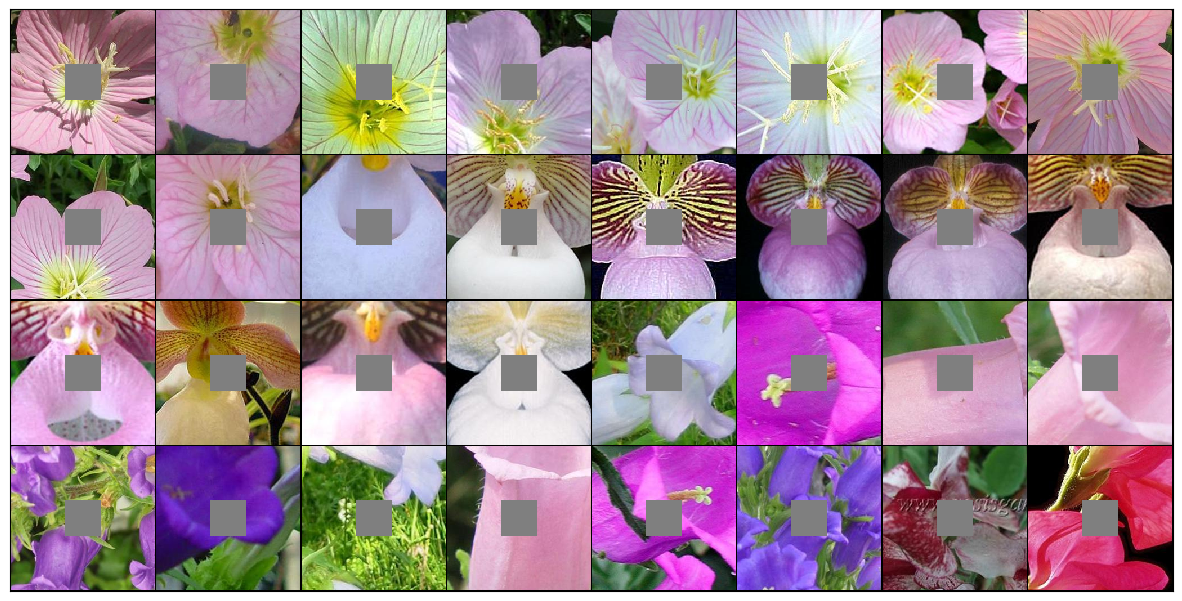}
    \caption{Masked Inputs.}
    \label{fig:masked-inputs}
\end{figure*}

\begin{figure*}
    \centering
    \includegraphics[width=\textwidth]{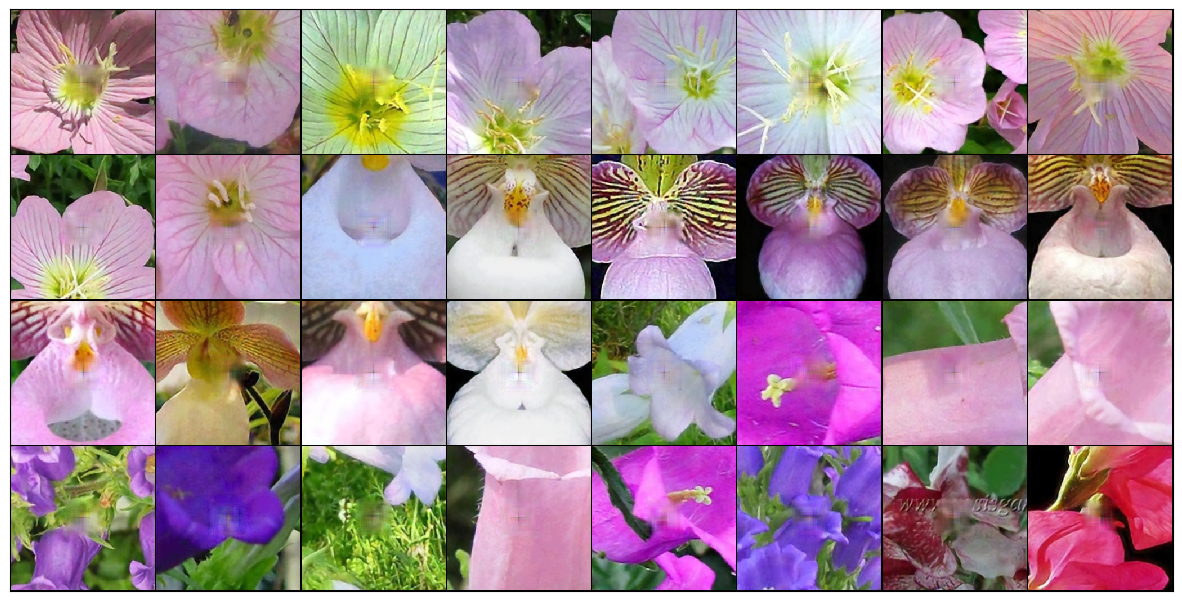}
    \caption{VQ-VAE Inpainting Results.}
    \label{fig:vqgan-inpainting}
\end{figure*}

\begin{figure*}
    \centering
    \includegraphics[width=\textwidth]{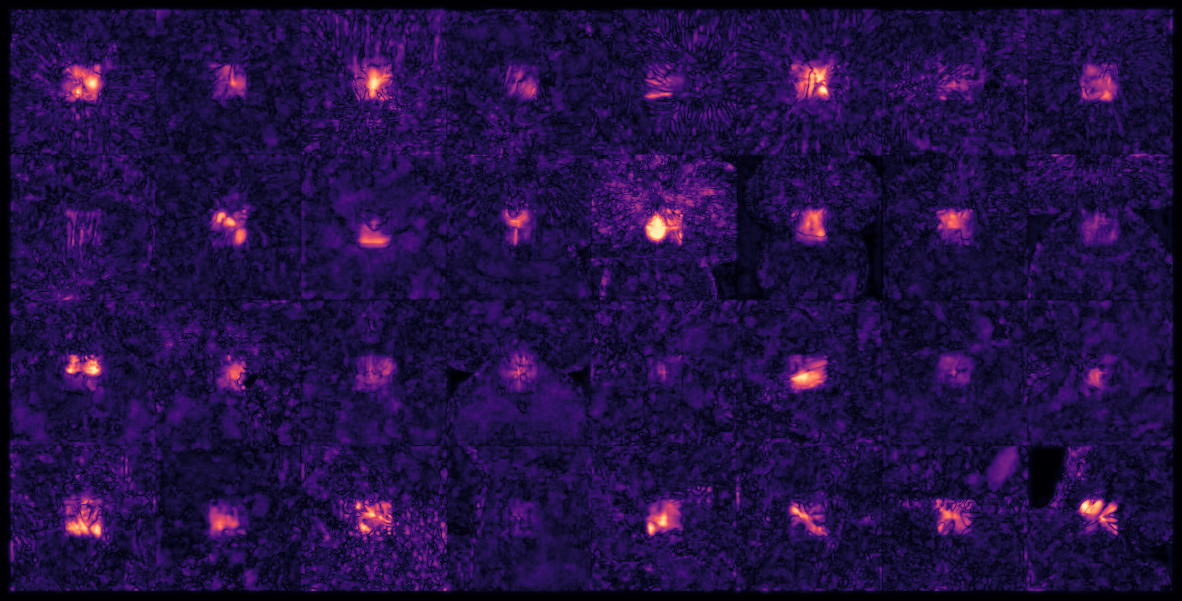}
    \caption{FLIP Heatmap Evaluations for VQ-VAE Inpainting.}
    \label{fig:vqgan-inpainting-flip}
\end{figure*}

\begin{figure*}
    \centering
    \includegraphics[width=\textwidth]{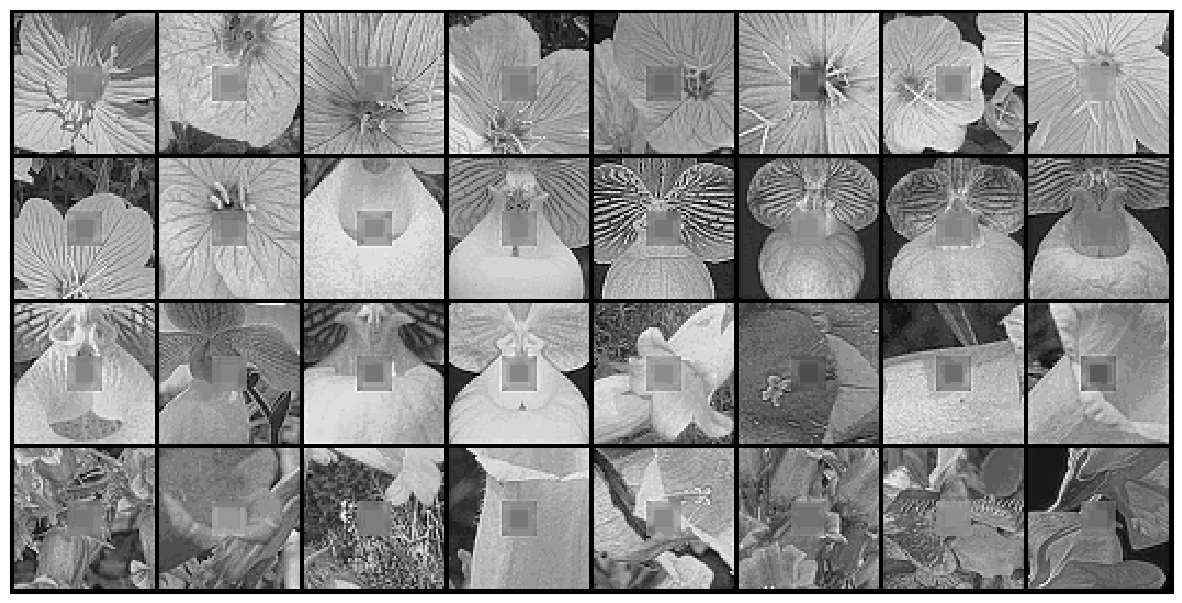}
    \caption{Top singular component of the VQ reconstructed latent space.}
    \label{fig:vqgan-inpainting-latents}
\end{figure*}

\begin{figure*}
    \centering
    \includegraphics[width=\textwidth]{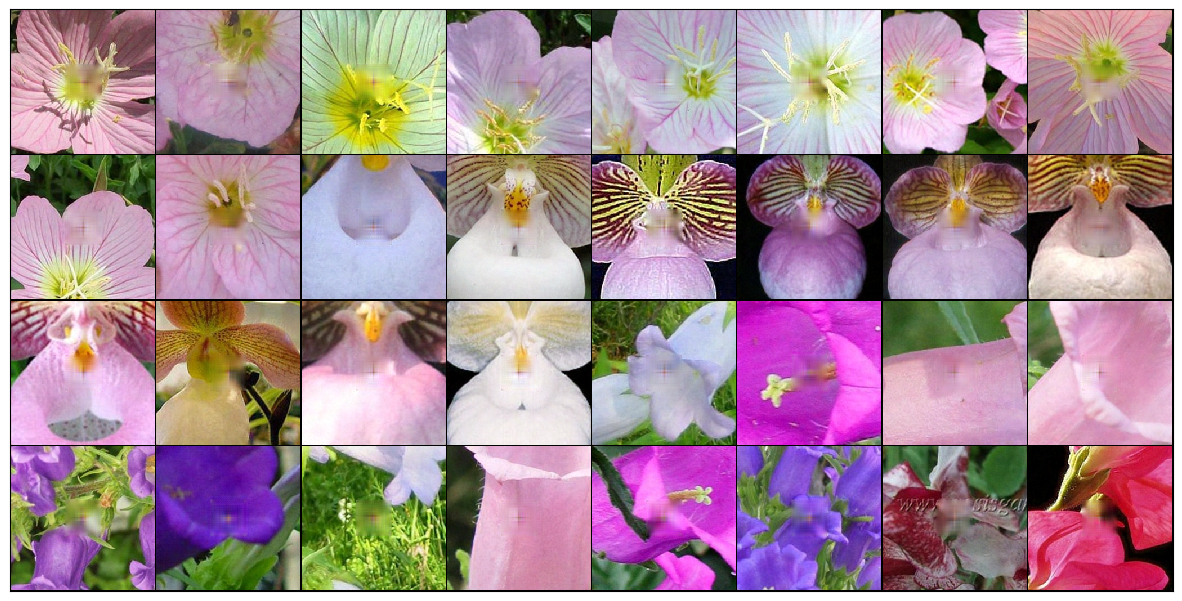}
    \caption{DL-VAE Inpainting Results.}
    \label{fig:dlgan-inpainting}
\end{figure*}

\begin{figure*}
    \centering
    \includegraphics[width=\textwidth]{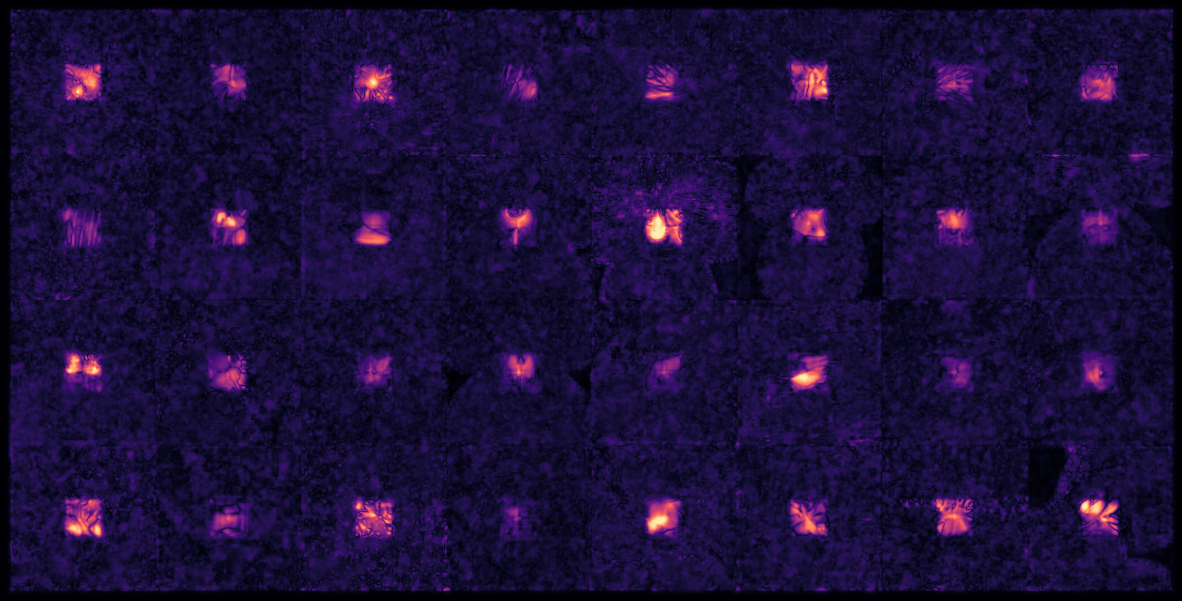}
    \caption{FLIP Heatmap Evaluations for DL-VAE Inpainting.}
    \label{fig:dlgan-inpainting-flip}
\end{figure*}

\begin{figure*}
    \centering
    \includegraphics[width=\textwidth]{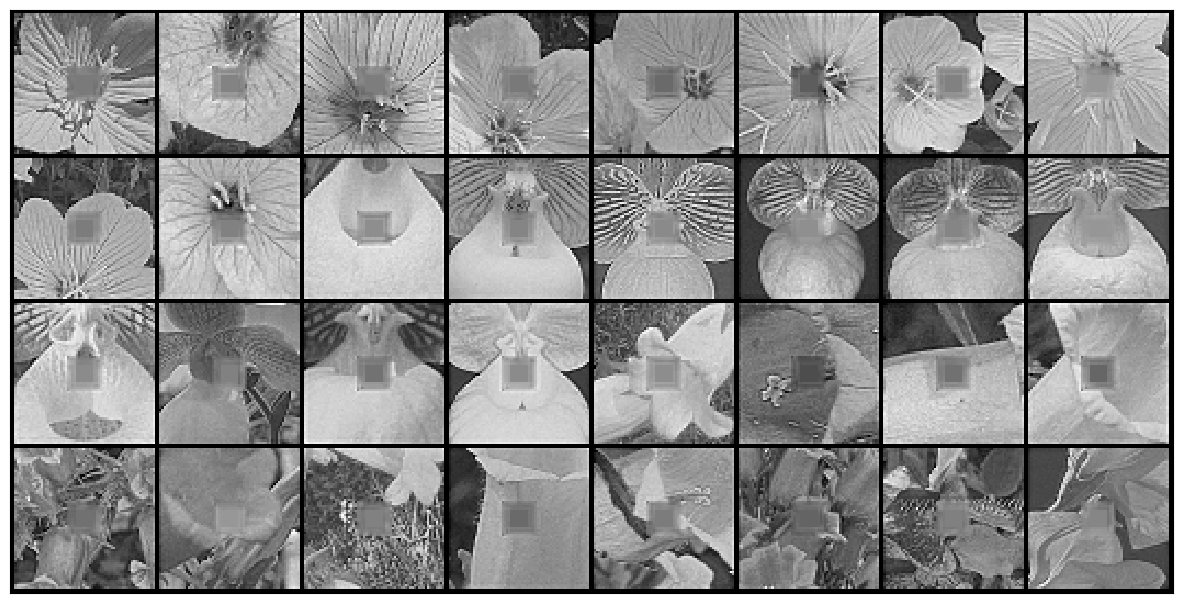}
    \caption{Top singular component of the DL reconstructed latent space.}
    \label{fig:dlgan-inpainting-latents}
\end{figure*}

The numerical evaluation results of the inpainting experiments are summarized in Table \ref{tab:inpainting}.

\begin{table}
  \centering
  {\small{
  \begin{tabular}{@{}lc@{}lc@{}@{}lc@{}lc@{}}
    \toprule
    Model &PSNR &\qquad FLIP &\quad FID\\
    \midrule
    VQ-VAE & $28.78$ &\quad\: $0.1301$ &\quad $72.19$\\
    DL-VAE & $\textbf{30.08}$ &\quad\: $\textbf{0.1098}$ &\quad $\textbf{59.48}$\\
    \bottomrule
  \end{tabular}
  }}
  \caption{Inpainting evaluations on the Oxford Flowers dataset.}
  \label{tab:inpainting}
\end{table}

We can observe interestingly although the DL-GAN still demonstrates better performance overall but is also sensitive to the added noise from masking as the VQ-VAE.

\medskip

\subsection{Restructuring the Latent Space of the Stable Diffusion Models}

We have also applied the VQ-VAE and DL-VAE models' compression bottlenecks to the latent space of the Stable Diffusion models \cite{ldm} for simple text-to-image generation tasks, during these experiments we fix the encoder and decoder network of the pretrained stable diffusion network along with the weights of the diffusion U-Net \cite{ldm, ddpm} and finetune the Vector Quantization bottleneck for a moderate amount of training steps ($100$), note that here a big advantage of our Dictionary Learning formulation is that once the dictionary is learned and fixed, since the sparse coding stage is deterministic, we do not need any additional finetuning to apply the Dictionary Learning bottleneck to the Stable Diffusion model. Some of the results from our experiments are shown in the images below,

\begin{figure*}
  \centering
  \begin{subfigure}{0.3\linewidth}
    \includegraphics[width=\textwidth]{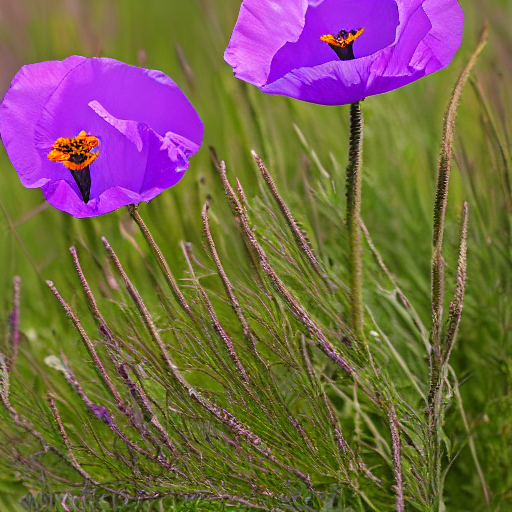}
    \caption{Original Encoder-Decoder Output.}
    \label{fig:ldm1}
  \end{subfigure}
  \hspace{1em}
  \begin{subfigure}{0.3\linewidth}
    \includegraphics[width=\textwidth]{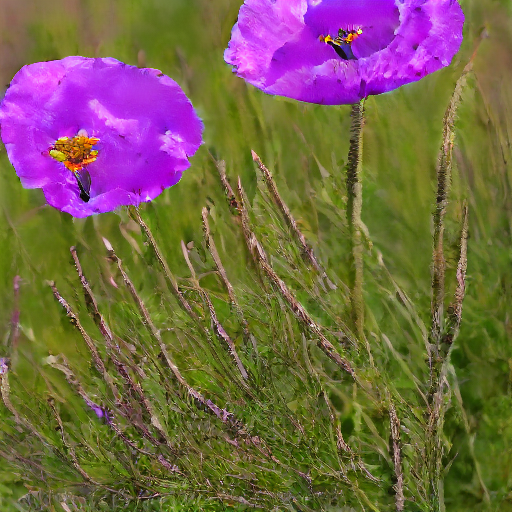}
    \caption{Output from Vector Quantized Latents.}
    \label{fig:vq-ldm1}
  \end{subfigure}
  \hspace{1em}
  \begin{subfigure}{0.3\linewidth}
    \includegraphics[width=\textwidth]{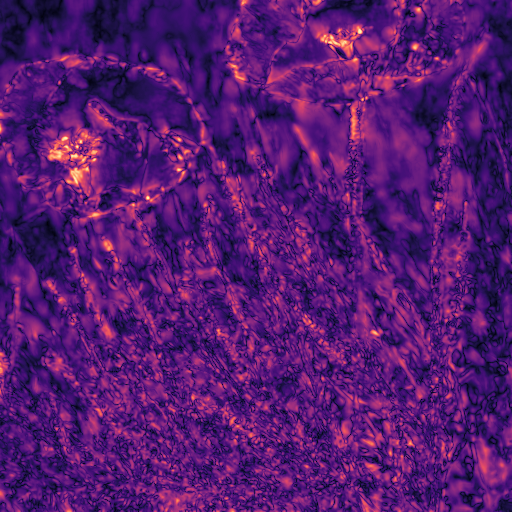}
    \caption{Flip Heatmap Evaluation.}
    \label{fig:vq-ldm1-flip}
  \end{subfigure}
  \caption{Comparison between the latent space attained from the original encoder-decoder of Stable Diffusion and the Stable Diffusion with vector quantized latent space given the Prompt: ``purple californian poppy'' (the flower is yellow).}
  \label{fig:vq-ldm1-results}
\end{figure*}

\begin{figure*}
  \centering
  \begin{subfigure}{0.3\linewidth}
    \includegraphics[width=\textwidth]{images/ldm1.png}
    \caption{Original Encoder-Decoder Output.}
  \end{subfigure}
  \hspace{1em}
  \begin{subfigure}{0.3\linewidth}
    \includegraphics[width=\textwidth]{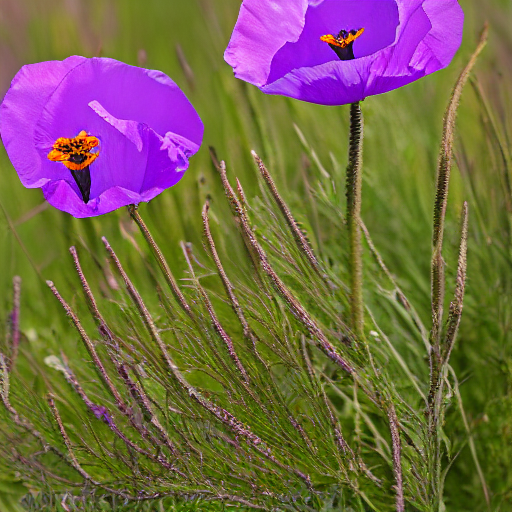}
    \caption{Output from Dictionary Learned Latents.}
    \label{fig:dl-ldm1}
  \end{subfigure}
  \hspace{1em}
  \begin{subfigure}{0.3\linewidth}
    \includegraphics[width=\textwidth]{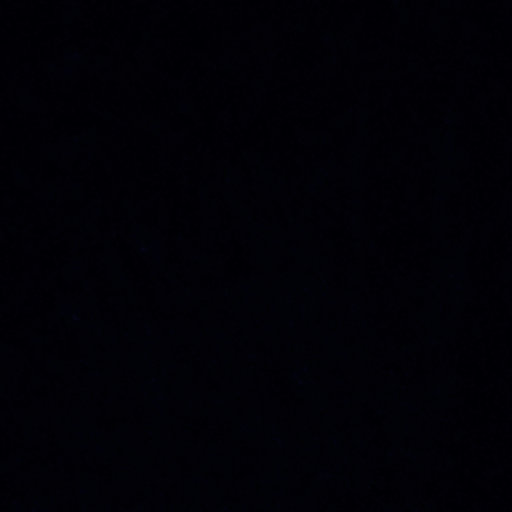}
    \caption{Flip Heatmap Evaluation.}
    \label{fig:dl-ldm1-flip}
  \end{subfigure}
  \caption{Comparison between the latent space attained from the original encoder-decoder of Stable Diffusion and the Stable Diffusion with dictionary learned latent space given the Prompt: ``purple californian poppy'' (the flower is yellow).}
  \label{fig:dl-ldm1-results}
\end{figure*}

\begin{figure*}
  \centering
  \begin{subfigure}{0.3\linewidth}
    \includegraphics[width=\textwidth]{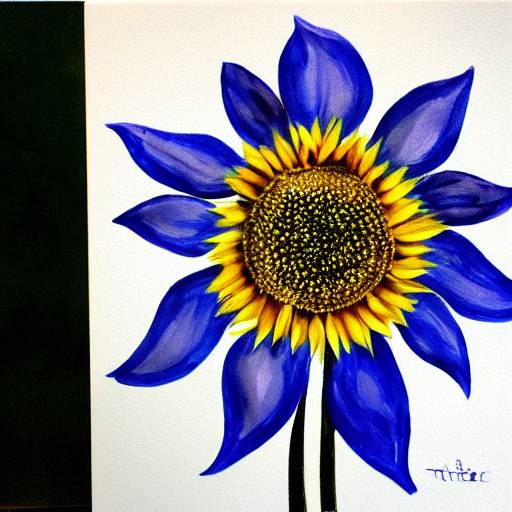}
    \caption{Original Encoder-Decoder Output.}
    \label{fig:ldm2}
  \end{subfigure}
  \hspace{1em}
  \begin{subfigure}{0.3\linewidth}
    \includegraphics[width=\textwidth]{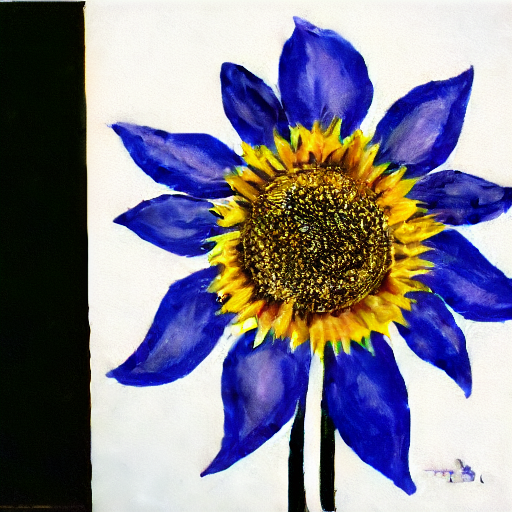}
    \caption{Output from Dictionary Learned Latents.}
    \label{fig:vq-ldm2}
  \end{subfigure}
  \hspace{1em}
  \begin{subfigure}{0.3\linewidth}
    \includegraphics[width=\textwidth]{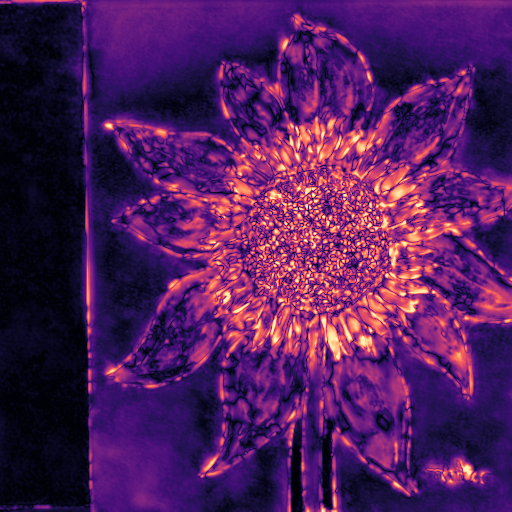}
    \caption{Flip Heatmap Evaluation.}
    \label{fig:vq-ldm2-flip}
  \end{subfigure}
  \caption{Comparison between the latent space attained from the original encoder-decoder of Stable Diffusion and the Stable Diffusion with vector quantized latent space given the Prompt: ``blue sunflower'' (the flower is yellow).}
  \label{fig:vq-ldm2-results}
\end{figure*}

\begin{figure*}
  \centering
  \begin{subfigure}{0.3\linewidth}
    \includegraphics[width=\textwidth]{images/ldm2.png}
    \caption{Original Encoder-Decoder Output.}
  \end{subfigure}
  \hspace{1em}
  \begin{subfigure}{0.3\linewidth}
    \includegraphics[width=\textwidth]{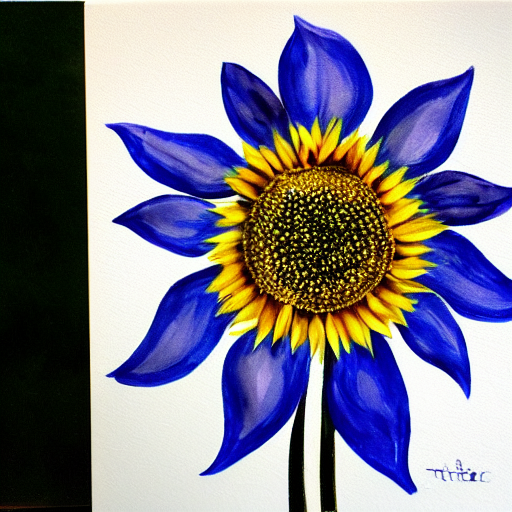}
    \caption{Output from Dictionary Learned Latents.}
    \label{fig:dl-ldm2}
  \end{subfigure}
  \hspace{1em}
  \begin{subfigure}{0.3\linewidth}
    \includegraphics[width=\textwidth]{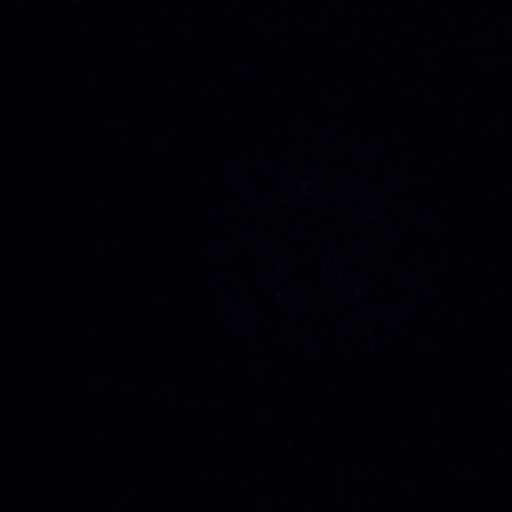}
    \caption{Flip Heatmap Evaluation.}
    \label{fig:dl-ldm2-flip}
  \end{subfigure}
  \caption{Comparison between the latent space attained from the original encoder-decoder of Stable Diffusion and the Stable Diffusion with dictionary learned latent space given the Prompt: ``blue sunflower'' (the flower is yellow).}
  \label{fig:dl-ldm2-results}
\end{figure*}

\begin{figure*}
  \centering
  \begin{subfigure}{0.3\linewidth}
    \includegraphics[width=\textwidth]{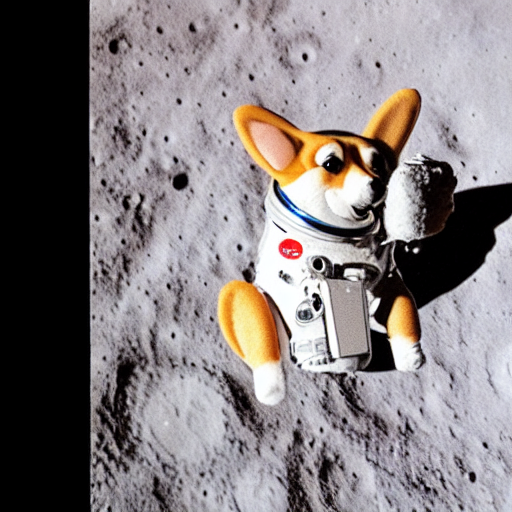}
    \caption{Original Encoder-Decoder Output.}
    \label{fig:ldm3}
  \end{subfigure}
  \hspace{1em}
  \begin{subfigure}{0.3\linewidth}
    \includegraphics[width=\textwidth]{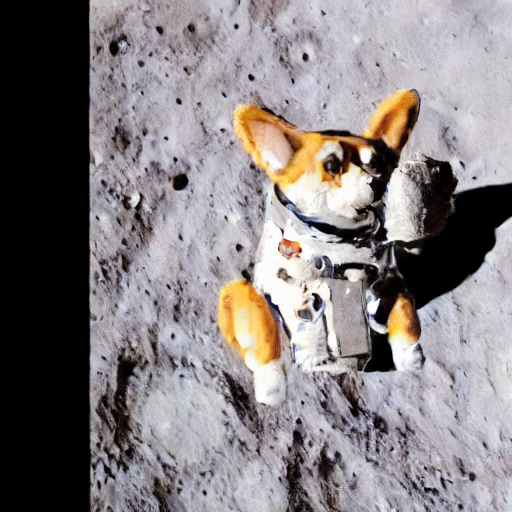}
    \caption{Output from Vector Quantized Latents.}
    \label{fig:vq-ldm3}
  \end{subfigure}
  \hspace{1em}
  \begin{subfigure}{0.3\linewidth}
    \includegraphics[width=\textwidth]{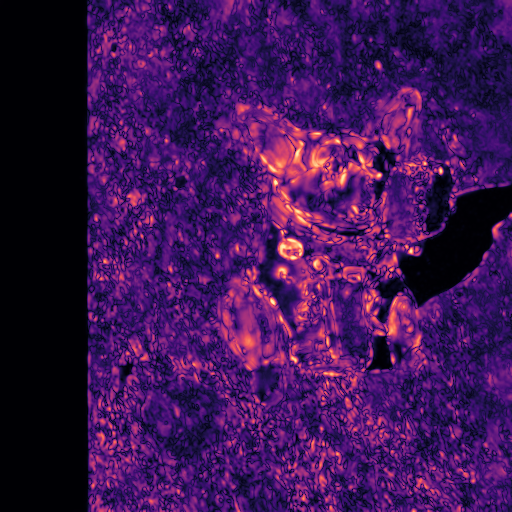}
    \caption{Flip Heatmap Evaluation.}
    \label{fig:vq-ldm3-flip}
  \end{subfigure}
  \caption{Comparison between the latent space attained from the original encoder-decoder of Stable Diffusion and the Stable Diffusion with vector quantized latent space given the Prompt: ``astronaut corgi on the moon''.}
  \label{fig:vq-ldm3-results}
\end{figure*}

\begin{figure*}
  \centering
  \begin{subfigure}{0.3\linewidth}
    \includegraphics[width=\textwidth]{images/ldm3.png}
    \caption{Original Encoder-Decoder Output.}
  \end{subfigure}
  \hspace{1em}
  \begin{subfigure}{0.3\linewidth}
    \includegraphics[width=\textwidth]{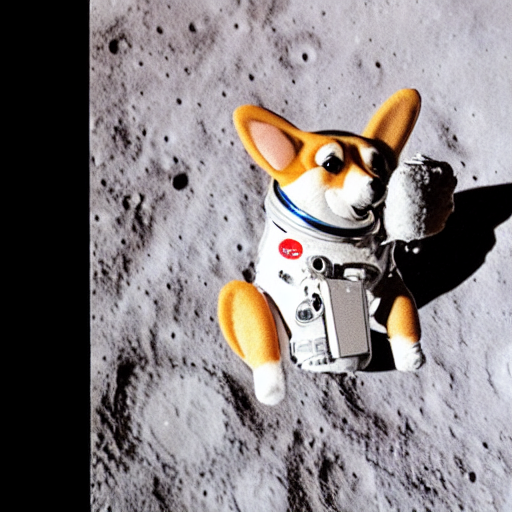}
    \caption{Output from Dictionary Learned Latents.}
    \label{fig:dl-ldm3}
  \end{subfigure}
  \hspace{1em}
  \begin{subfigure}{0.3\linewidth}
    \includegraphics[width=\textwidth]{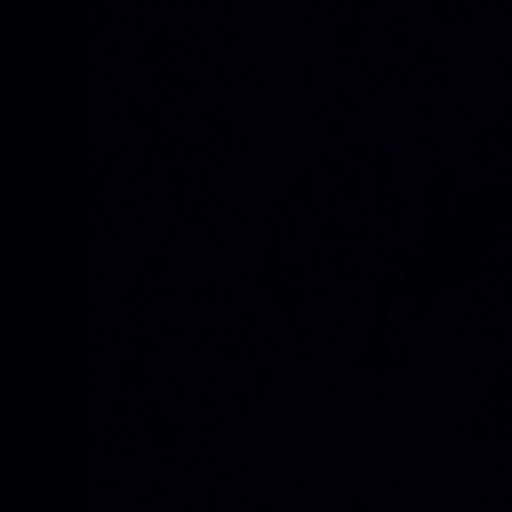}
    \caption{Flip Heatmap Evaluation.}
    \label{fig:dl-ldm3-flip}
  \end{subfigure}
  \caption{Comparison between the latent space attained from the original encoder-decoder of Stable Diffusion and the Stable Diffusion with dictionary learned latent space given the Prompt: ``astronaut corgi on the moon''.}
  \label{fig:dl-ldm3-results}
\end{figure*}

The numerical evaluation results of the inpainting experiments are summarized in Table \ref{tab:ldm}.

\begin{table}
  \centering
  {\small{
  \begin{tabular}{@{}lc@{}lc@{}@{}lc@{}lc@{}}
    \toprule
    Structure &PSNR &\qquad FLIP &\quad FID\\
    \midrule
    Vector Quantization & $14.70$ &\quad\: $0.1301$ &\quad $63.90$\\
    Dictionary Learning & $\textbf{58.06}$ &\quad\: $\textbf{0.006690}$ &\quad $\textbf{0.01061}$\\
    \bottomrule
  \end{tabular}
  }}
  \caption{Evaluations on the Latent Space Structuring Strategies for Stable Diffusion.}
  \label{tab:ldm}
\end{table}

From the experiments we can clearly see the quantization artifacts from the Vector Quantization bottleneck, the Dictionary Leaning Bottleneck, on the other hand, reconstructs the latent space with high fidelity.

\end{document}